\documentclass[conference]{IEEEtran}
\IEEEoverridecommandlockouts
\usepackage{cite}
\usepackage{amsmath,amssymb,amsfonts}
\usepackage{amsthm}
\usepackage{graphicx}
\usepackage{textcomp}
\usepackage{xcolor}
\usepackage{algorithm,algcompatible,amsmath}

\usepackage{mathtools}
\usepackage{thmtools} 
\usepackage{booktabs}
\usepackage{hyperref}

\usepackage{comment}

\newtheorem{definition}{Definition}

\usepackage{graphicx}
\usepackage{textcomp}
\usepackage{xcolor}
\def\BibTeX{{\rm B\kern-.05em{\sc i\kern-.025em b}\kern-.08em
    T\kern-.1667em\lower.7ex\hbox{E}\kern-.125emX}}
\begin{document}

\title{Robust Time Series Chain Discovery with Incremental Nearest Neighbors\\
}
\author{\IEEEauthorblockN{Li Zhang\IEEEauthorrefmark{1}\IEEEauthorrefmark{4}, Yan Zhu\IEEEauthorrefmark{2}\IEEEauthorrefmark{4}, Yifeng Gao\IEEEauthorrefmark{3}, Jessica Lin\IEEEauthorrefmark{1}}
\IEEEauthorblockA{
\textit{\IEEEauthorrefmark{1}Department of Computer Science,George Mason University, USA}\\
\textit{\IEEEauthorrefmark{2}Google, USA}\\
\textit{\IEEEauthorrefmark{3}Department of Computer Science,University of Texas Rio Grande Valley, USA}\\
\{lzhang18,jessica\}@gmu.edu, zhuyan@google.com, yifeng.gao@utrgv.edu}
\IEEEcompsocitemizethanks{\IEEEcompsocthanksitem\IEEEauthorrefmark{4}equal contribution}
\vspace{-10mm}
}

\maketitle

\begin{abstract}
  Time series motif discovery has been a fundamental task to identify meaningful repeated patterns in time series. Recently, time series chains were introduced as an expansion of time series motifs to identify the continuous evolving patterns in time series data. Informally, a time series chain (TSC) is a temporally ordered set of time series subsequences, in which every subsequence is similar to the one that precedes it, but the last and the first can be arbitrarily dissimilar. TSCs are shown to be able to reveal latent continuous evolving trends in the time series, and identify precursors of unusual events in complex systems. Despite its promising interpretability, unfortunately, we have observed that existing TSC definitions lack the ability to accurately cover the evolving part of a time series: the discovered chains can be easily cut by noise and can include non-evolving patterns, making them impractical in real-world applications. Inspired by a recent work that tracks how the nearest neighbor of a time series subsequence changes over time, we introduce a new TSC definition which is much more robust to noise in the data, in the sense that they can better locate the evolving patterns while excluding the non-evolving ones. We further propose two new quality metrics to rank the discovered chains. With extensive empirical evaluations, we demonstrate that the proposed TSC definition is significantly more robust to noise than the state of the art, and the top ranked chains discovered can reveal meaningful regularities in a variety of real world datasets.

\end{abstract}

\begin{IEEEkeywords}
time series, motifs, chains, drift, prognostics
\end{IEEEkeywords}

\section{Introduction}

In the last two decades, the task of finding repetitive patterns in time series data, known as motif discovery, has received a lot of attentions in the research community due to its wide range of applications across many different domains~\cite{zhumatrix, zhang2020semantic, yeh2016matrix,gao2017trajviz}. Recently, a new primitive called the time series chain (TSC) is introduced as a new tool to capture the \textit{evolving} patterns in the data over time \cite{zhu2017matrix,imamura2020matrix}. Informally, a time series chain is an ordered set of subsequences extracted from a time series, where adjacent subsequences in the chain are similar, but the first and the last are arbitrarily dissimilar. Different from other time series data mining tasks such as motif discovery~\cite{zhumatrix, yeh2016matrix, gao2018exploring}, discord discovery~\cite{keogh2005hot,senin2015time}, time series clustering~\cite{paparrizos2015k, ma2019learning, li2021time, anton2018time}, etc.,
time series chains can capture any potential \textit{drift} accumulated over time, which widely exists in many complex systems, natural phenomena and societal changes~\cite{bloch1997practical,zhu2017matrix,imamura2020matrix}.

\begin{figure}
    \centering
    \includegraphics [width=85mm,height=5cm,
  keepaspectratio]{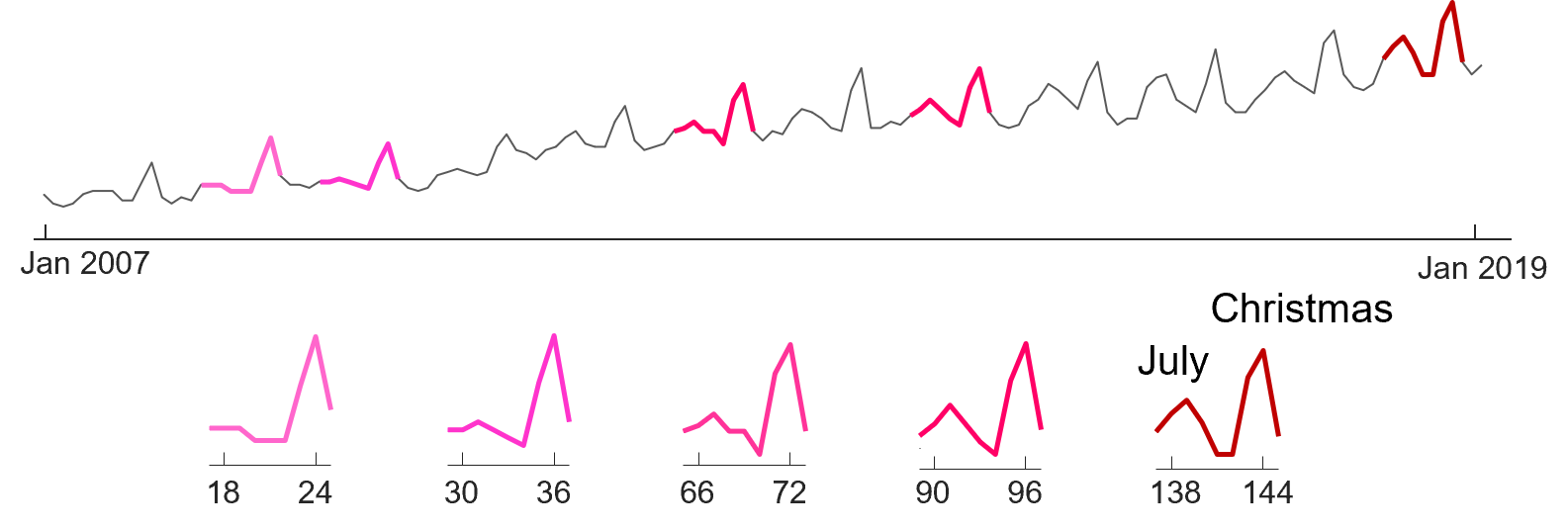}
    \caption{The time series chain found in the 12-year Google search volume for query "Costco" indicates the growing importance of the July 4\textsuperscript{th} sales event.}
    \label{fig:costcogoogletrend}
\end{figure}

To help the reader better understand the capability of time series chains, consider Google Trend data in Fig. \ref{fig:costcogoogletrend} corresponding to the search term ``Costco", an American retail chain over the course of 12 years. The time series chain discovered is highlighted in the figure. The first pattern in the chain shows a sharp peak around Christmas, which is normal due to the increased online shopping needs around the holiday season. Then as that peak becomes smoother and smoother over the years, a small bump emerges at around July, which gradually grows into another major peak in the pattern. The evolving patterns in the chain clearly reveal people's growing interest in Costco's 4\textsuperscript{th} of July online sales event, which provides useful marketing insights.

Another typical application of time series chains is prognostics.
As noted by previous studies~\cite{ahmad2012overview, bloch1997practical}, ``\textit{most equipment failures are preceded by certain signs, conditions, or indications that such a failure was going to occur.}'' Time series chains can identify not only a single precursor, but also a whole sequence of patterns revealing the \textit{continuous and gradual} change of the system, helping analyzers uncover the reasons at an early stage and prevent catastrophic failures.

So far there are two methods to discovery time series chains. The concept of chains was first introduced in \cite{zhu2017matrix} based on a bi-directional definition, and later on \cite{imamura2020matrix} proposed a geometric definition to improve robustness. Following~\cite{imamura2020matrix}, for the rest of the paper we refer to the original time series chain definition in \cite{zhu2017matrix} as TSC17, and the later geometric chain definition in \cite{imamura2020matrix} as TSC20.

\begin{figure}[t]
    \centering
    \includegraphics [width=85mm,height=5cm,
  keepaspectratio]{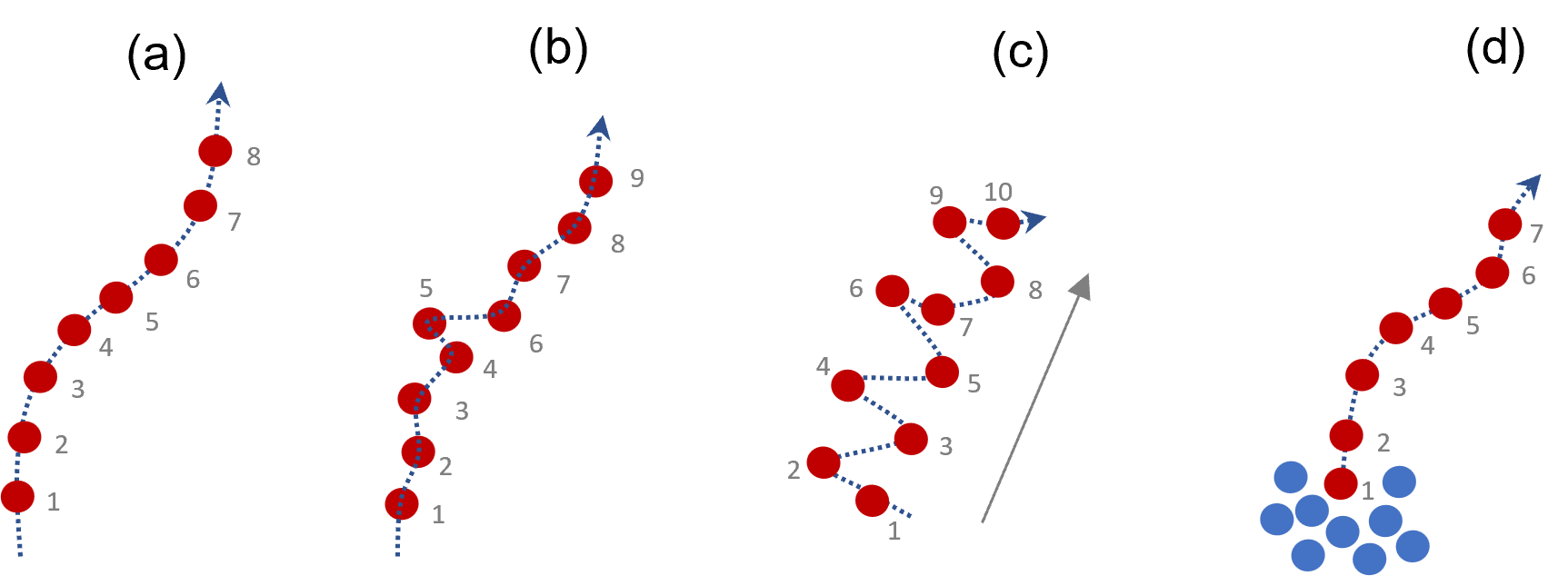}
    \caption{Example chains in the two-dimensional space. (a) A clean chain that can be discovered by both TSC17 and TSC20. (b) A chain with a single deviated node (Node 5) that would be missed by TSC17. (c) A chain that contains more fluctuations but still shows an evolving trend. This chain cannot be found by either TSC17 or TSC20. (d) A chain (the red nodes) drifting from a steady state (the blue nodes). TSC20 would attach the blue nodes to the discovered chain, which is undesirable.}
    \label{fig:introexample}
    \vspace{-6mm}
\end{figure}

Despite the superior interpretability of time series chains, unfortunately, we have observed that existing chain discovery methods only work well in relatively ideal situations when the data is clean and the patterns evolve in small, quasi-linear steps (as shown in Fig. \ref{fig:introexample}.a). In practice, the continuous data collection process are often affected by environment changes, human errors, sensor signal interruption, etc. \cite{jardine2006review}. As a result, the data is noisy: some patterns can deviate from the general trend (e.g. Node 5 in Fig. \ref{fig:introexample}.b), consecutive patterns can be flipped, the chain can evolve in a zigzag fashion due to fluctuations in data (Fig. \ref{fig:introexample}.c). It is also possible that the patterns remain relatively stable before they start to drift (which is typical in prognostics, as shown in Fig. \ref{fig:introexample}.d). As we will later demonstrate in Section VI, existing chain discovery methods are not robust enough to handle such scenarios in real-world data.

Concretely, we found the following limitations in existing time series chain definitions:
\begin{itemize}
    \item The constraints introduced in TSC17 are too strict in that a small fluctuation can easily `cut' a chain. Fig. \ref{fig:introexample}.b shows an example that does not satisfy the constraints. 
    
    \item The direction angle threshold introduced in TSC20 is very sensitive to noise. When the data contain fluctuations, the definition can miss an apparent chain (e.g., Fig. \ref{fig:introexample}.c) even with a 50 degree threshold (the recommended threshold is 40 degrees). However, if we further increase the threshold, the discovered chains would include a lot of noisy patterns unrelated to the evolving trend.
    
    \item Even with a very small direction angle, chains found by TSC20 can include patterns unrelated to the evolving trend. For example, in the case when the data is first stable (where the patterns are very similar) and then start to drift (e.g., Fig. \ref{fig:introexample}.d), TSC20 will attach the stable (blue) patterns to the evolving chain. As a result, it is unable to tell \textit{when} the system starts to drift.
\end{itemize}

To address these limitations, in this work we propose a novel time series chain definition $TSC22$, which exploits an idea to track how the nearest neighbor of a pattern changes over time to improve the robustness of chain discovery. Similar to $TSC17$, $TSC22$ is a bi-directional definition but with much more relaxed constraints, and it does not rely on any angle constraint as in $TSC20$. Furthermore, we propose a new quality metric to rank the chains discovered by our method. With extensive empirical evaluations, we demonstrate that the proposed method is much more robust to noise than the state of the art, and the top ranked chains discovered can reveal meaningful regularities in a variety of real world datasets.

The rest of the paper is organized as follows. In Section II we briefly review the background and related work. Section III goes over basic time series notations and existing time series chain definitions. Section IV shows the theoretical limitations of the existing definitions. In Section V we introduce our new time series chain definition as well as the ranking criteria. Section VI demonstrates the effectiveness of our proposed method through extensive evaluations on both real-world and synthetic data. Section VII concludes the paper.
\section{BACKGROUND AND RELATED WORK}

 The review is brief as the time series chain is a relatively new topic. The most closely related works are the ones by Zhu et al.~\cite{zhu2017matrix} and Imamura et al.~\cite{imamura2020matrix}. Zhu et al.~\cite{zhu2017matrix} first introduce the concept of time series chains. The work enforces every adjacent pair of subsequences in the chain to be the left and right nearest neighbors of each other, and reports the longest chain as the top chain in the time series. Although the concept is simple and intuitive, the bi-directional nearest neighbor requirement is shown to be too strict in many real-world applications, as the chain can easily break by data fluctuations and noise \cite{imamura2020matrix}. Imamura et al.~\cite{imamura2020matrix} relax the bi-direction condition by only enforcing the single-directional left nearest neighbor requirement, and added a pre-defined angle constraint to guarantee the directionality of the discovered chains. The concept is shown to be more robust than \cite{zhu2017matrix} in some conditions, but a new angle parameter is introduced, and we have observed that in some real-world scenarios, no meaningful chains can be found no matter how we set this parameter. We will elaborate more on the theoretical limitations of these two chain definitions in Section IV. 

Other works on time series chains have different problem settings. Wang et al.~\cite{wang2019discovering} explores methods to speed up chain discovery in streaming data using the existing time series chain definition~\cite{zhu2017matrix}. Zhang et al.~\cite{zhang2022joint} propose a method to detect joint time series chains across two time series, while we focus on improving the chain definition in a single time series. 

\section{DEFINITIONS}\label{sec:definition}
In this section, we first review necessary time series notations, then consider the formal definition of time series chains.
\subsection{Time Series Notations}
We start with fundamental definitions related to time series.
\vspace{-4mm}
\begin{definition}
A \textbf{time series} $T = [t_1, t_2, \ldots, t_n]$ is an ordered list of data points, where $t_i$ is a finite real number and $n$ is the length of time series $ T$.
\end{definition}
\begin{definition}
A \textbf{time series subsequence} $S^T_i=[t_i, t_{i+1}, \ldots, t_{i+l-1}]$ is a contiguous set of points in time series $T$ starting from position $i$ with length $l$. Typically $l \ll n$, and $1\leq i\leq n-l+1$.
\end{definition}
For simplicity, if we only consider only one time series, we use $S_i$ interchangably with $S^T_i$. 
\noindent Subsequences can be extracted from time series $T$ by sliding a  window of fixed length $l$ across the time series. Given a query subsequence, we can compute its distance to every subsequence in a time series $T$. This is called a \emph{distance profile}~\cite{yeh2016matrix}:
\vspace{-1mm}
\begin{definition}
A \textbf{distance profile} $D_{Q,T}$ is a vector containing the distances between a query subsequence $Q$ and every subsequence of the same length in time series $T$. Formally, $D_{Q,T} = [d(Q,S_1), d(Q,S_2), \ldots, d(Q,S_{n-l+1})]$, where $d(.,.)$ denotes a distance function. In the special case where $Q$ is a subsequence of time series $T$ starting at position $i$, we denote the distance profile as $D_i$.
\end{definition}

Following the original Matrix Profile works~\cite{yeh2016matrix, zhumatrix}, here we use the z-normalized Euclidean distance instead of the Euclidean distance to achieve scale and offset invariance~\cite{lin2007experiencing}. 
Note that comparing a subsequence to itself or those that largely overlap with it is not meaningful. As such, we avoid comparing subsequences if they overlap with each other by more than $l/2$ where $l$ is the subsequence length.

We can further divide the distance profile into a \emph{left distance profile} and a \emph{right distance profile}~\cite{zhu2017matrix}.
\begin{definition}
A \textbf{left distance profile} $DL_i$ of time series $T$ is a vector containing the Euclidean distances between a given subsequence $S_i \in T$ and every subsequence to the left of $S_i$. Formally, $DL_i = [d(S_i,S_1), d(S_i,S_2), \ldots, d(S_i,S_{i-l/2})].$
\end{definition}

\begin{definition}
A \textbf{right distance profile} $DR_i$ is a vector where $DR_i = [d(S_i,S_{i+l/2}),\ldots, d(S_i,S_{n-l+1})]$.
\end{definition}
\vspace{-1mm}

We can easily find the \textbf{\textit{left nearest neighbor (LNN)}} and the \textbf{\textit{right nearest neighbor (RNN)}} of a subsequence by examining the minimum values in the left and right distance profiles respectively. We use two vectors, the \emph{left matrix profile} and the \emph{right matrix profile} to store the distances between all subsequences and their corresponding left/right nearest neighbors, as well as the location of these nearest neighbors.
\vspace{-1mm}
\begin{definition}
A \textbf{left matrix profile} $MP_L$ is a two dimensional vector of size $2\times (n-l+1)$ where $MP_L(1,i)=\min(DL_i)$ and $MP_L(2,i)=\arg\min(DL_i)$ 
\end{definition}
Here $MP_L(1,i)$ stores the distances between $S_i$ and the subsequence most similar to it \emph{before} $i$ (i.e., its LNN), and $MP_L(2,i)$ 
stores the location of the $LNN$. 
\vspace{-1mm}
\begin{definition}
A \textbf{right matrix profile} $MP_R$ is a two dimensional vector of size $2\times (n-l+1)$ where $MP_R(1,i)=\min(DR_i)$ and $MP_R(2,i)=\arg\min(DR_i)$.
\end{definition}
\vspace{-2mm}
\subsection{Existing Time Series Chain Definitions}
Existing time series chain definitions \cite{imamura2020matrix}\cite{zhu2017matrix} are both developed upon the \emph{backward time sereis chain}, which can be easily obtained given the left matrix profile.
\begin{definition}
A \textbf{backward chain} $TSC^{BWD}$ of time series $T$ is a finite ordered set of time series subsequences: $TSC^{BWD} =[S_{C_1}, S_{C_2}, S_{C_3}, \ldots, S_{C_m}]$, where $C_1 > C_2 >\ldots > C_m$ are indices in time series $T$, such that  for any $1\leq i< m$, we have $LNN(S_{C_{i}}) = S_{C_i+1}$. 
\end{definition}
\vspace{-1mm}
For clarity, we denote a subsequence in a time series chain as a \textbf{node}. We call the first node in the backward chain the \textbf{start node} and the last node the \textbf{end node}. Here in $TSC^{BWD}$, $S_{C_1}$ is start node and $S_{C_m}$ is the end node. For example, if the subsequence length is 1, in a backward chain $[10, 5, 2, 1]$, the start node would be 10 and the end node would be 1. 

\begin{figure}
    \centering
    \includegraphics [width=75mm,height=3cm,
  keepaspectratio]{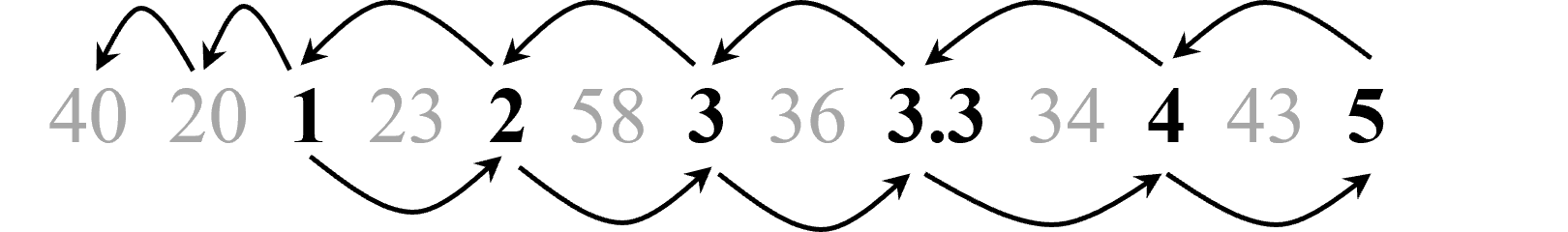}
   \vspace{-3mm}
    \caption{TSC17 looks for bi-directional connections between susbsequences. A left pointing arrow links a subsequence to its left nearest neighbor ($LNN$), while a right pointing arrow links a subsequence to its right nearest neighbor ($RNN$). Here the chain breaks at 1, as $LNN(1)=20$ but $RNN(20)=23\neq1$.}
    \label{fig:TSC17definition}
\end{figure}

Analogously, we can obtain the \emph{forward time series chain} from the right matrix profile. 
\vspace{-1mm}
\begin{definition}
A \textbf{forward chain} $TSC^{FWD}$ of time series $T$ is a finite ordered set of time series subsequences:  $TSC^{FWD} =[S_{C_1}, S_{C_2}, S_{C_3}, \ldots, S_{C_m}]$ where $C_1 < C_2 < \ldots < C_m$ are indices in time series $T$, such that for any $1\leq i<m$, we have $RNN(S_{C_{i}}) = S_{C_{i+1}}$. 
\end{definition}
\vspace{-1mm}
Existing works \cite{zhu2017matrix} \cite{imamura2020matrix} define chains by placing different constraints on the the backward chain. In \cite{zhu2017matrix}, a time series chain is defined based on the intersection of a backward chain and a forward chain. We call this a \emph{bi-directional time series chain}:
\vspace{-1mm}
\begin{definition}
A \textbf{bi-directional time series chain (TSC17)} of a time series $T$ is a finite ordered set of time series subsequences: $TSC =[S_{C_1}, S_{C_2}, S_{C_3}, \ldots, S_{C_m}]$ where $C_1>C_2 >\ldots>C_m$, such that for any $1\leq i<m$, we have $LNN(S_{C_{i}}) = S_{C_i+1}$ and $RNN(S_{C_{i+1}}) = S_{C_{i}}$.
\label{def:tsc17}
\end{definition}
\vspace{-1mm}
Here we denote the size of the set $m$ as the \textbf{length} of the chain.
To help the reader understand how TSC17 works, consider the time series in Fig. \ref{fig:TSC17definition}:
\vspace{-1mm}
\begin{center}
40,   20,   1,   23,   2,   58,   3,   36,   3.3,   34,   4,   43,   5

\end{center}

\vspace{-1mm}
Assume that the subsequence length is 1, and we use the absolute difference between the numbers to measure their distance. Following the backward chain $5\rightarrow4\rightarrow3.3\rightarrow3\rightarrow2\rightarrow1\rightarrow20\rightarrow40$, we check whether the LNN and the RNN of each node can form a loop:
\vspace{-2mm}
\begin{align*}
    LNN(5)=4 & \text{ and }RNN(4)=5;\\ 
    LNN(4)=3.3 & \text{ and } RNN(3.3)=4;\\ 
    LNN(3.3)=3 & \text{ and } RNN(3)=3.3;\\ 
    LNN(3)=2 & \text{ and } RNN(2)=3; \\ 
    LNN(2)=1 & \text{ and } RNN(1)=2.
\end{align*}
Since $LNN(1)=20$, but $RNN(20)=23\neq1$, the chain breaks at $1$. As shown in Fig. \ref{fig:TSC17definition}, the extracted chain (going backward) is:
\vspace{-2mm}
\[
    5\leftrightarrow  4 \leftrightarrow 3.3 \leftrightarrow 3 \leftrightarrow 2 \leftrightarrow 1
\vspace{-1mm}
\]

We can see that the numbers in the chain are gradually decreasing. A diligent reader might wonder why we cannot use the backward chain directly, but instead need to place extra constraints on it. 

A closer look at Fig. \ref{fig:TSC17definition} provides a good answer to this question. If we simply follow the backward chain $5\rightarrow 4 \rightarrow 3.3 \rightarrow 3 \rightarrow 2 \rightarrow 1 \rightarrow 20 \rightarrow 40$ without any extra constraint, then the chain will lose its "\textit{directionality}": the noisy big numbers (20, 40) simply do not conform to the slowly decreasing trend. The bi-directional constraint on TSC17 removes these noisy signals from the chain.

In \cite{imamura2020matrix}, a time series chain is defined by placing an angle constraint on the backward chain. We call this a \emph{geometric time series chain}:
\vspace{-2mm}
\begin{definition}
A \textbf{geometric time series chain (TSC20)} of a time series $T$ is a finite ordered set of time series subsequences: $TSC =[S_{C_1}, S_{C_2}, S_{C_3}, \ldots, S_{C_m}]$ $(C_1>C_2 > \ldots>C_m)$, such that for any $1\leq i<m$, we have $LNN(S_{C_{i}}) = S_{C_{i+1}}$, and for any $2\leq i<m$, we have the $i$-th angle $\theta_{i}\leq \theta$, where
$\theta_i = cos^{-1}\langle \frac{S_{C_i}-S_{C_1}}{\|S_{C_i}-S_{C_1}\|}, \frac{S_{C_{i+1}}-S_{C_1}}{\|S_{C_{i+1}}-S_{C_1}\|} \rangle$,
\noindent and $\theta$ is a predefined threshold.
\end{definition}

\begin{figure}
    \centering
    \vspace{-3mm}
    \includegraphics [width=80mm,height=3cm,
  keepaspectratio]{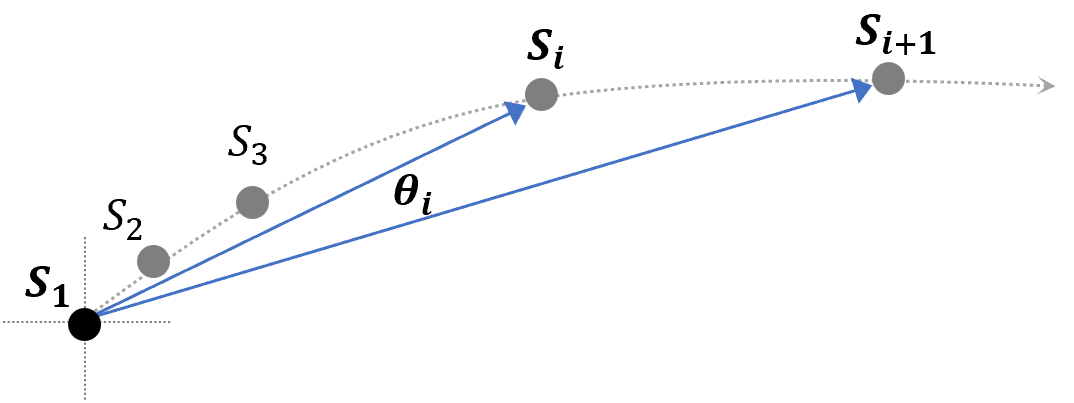}
    \caption{TSC20 places an angle constraint on the backward chain.}
    \label{fig:TSC20definition}
\end{figure}

Fig. \ref{fig:TSC20definition} illustrates how the TSC20 definition works in the 2-dimensional space. The direction angle $\theta_{i}$ measures the direction change from $S_i$ to $S_{i+1}$ based on the anchor node $S_1$. A small angle threshold between every two consecutive subsequences in the chain ensures that the subsequences evolve in similar directions.

\section{Limitations of TSC17 and TSC20}
Although both TSC17 and TSC20 are intuitive, we find them both vulnerable to noise and fluctuations in the data.

\subsection{The weakness of TSC17}
\begin{figure}
    \centering
    \includegraphics [width=80mm,height=3cm,
  keepaspectratio]{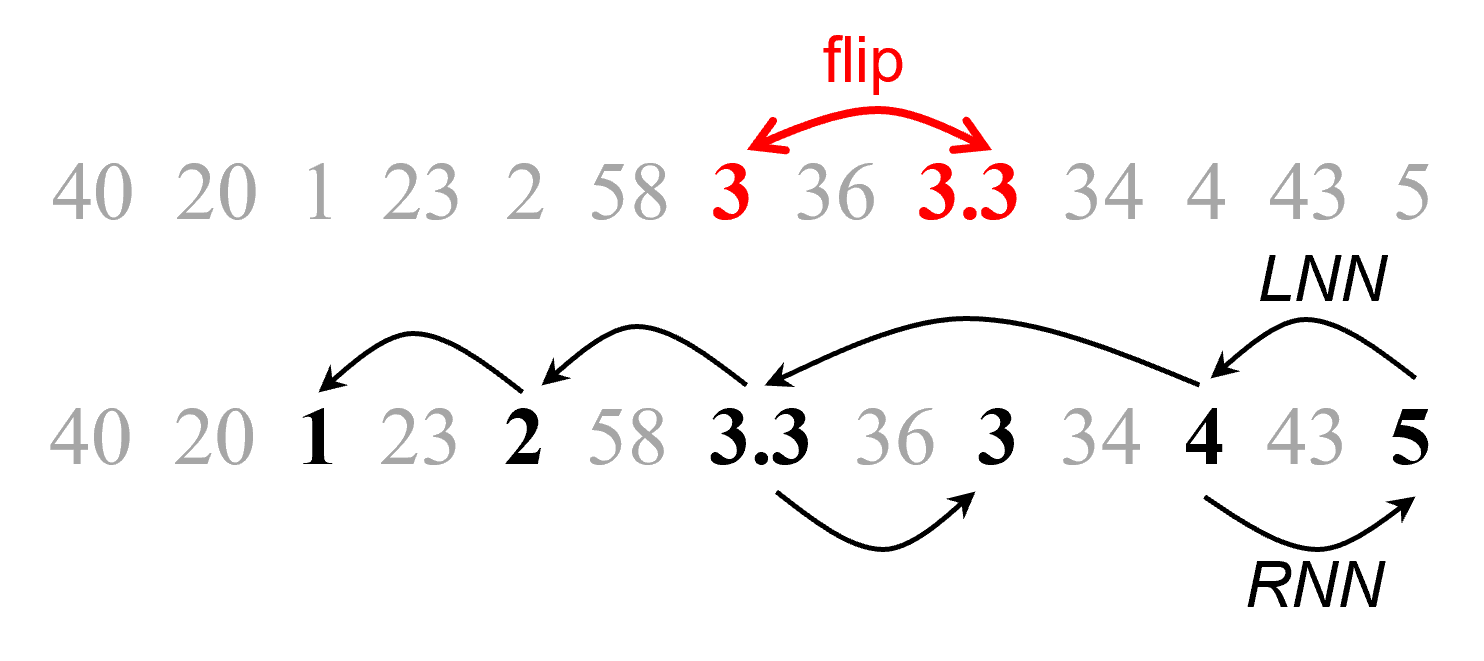}
    \caption{A flip of the subsequences can easily break a bi-directional chain (TSC17).}
    \label{fig:TSC17limitation}
\end{figure}

Let us revisit the example in Fig. \ref{fig:TSC17definition}. Suppose we flip the numbers 3 and 3.3, as shown in Fig. \ref{fig:TSC17limitation}.top. Now the time series becomes:
\vspace{-1mm}
\begin{center}
40,   20,   1,   23,   2,   58,   3.3,   36,   3,  34,   4,   43,   5
\end{center}
Consider growing a chain backward from node 5, as shown in Fig.~\ref{fig:TSC17limitation}.bottom. We can see that $LNN(5)=4$, $LNN(4)=5$, so we can add $4$ to the chain. Continuing with the process, $LNN(4)=3.3$, but $RNN(3.3)=3\neq4$, so 3.3 cannot be added and the chain breaks. However, from Fig. \ref{fig:TSC17limitation}.bottom we can see that even after flipping, the sequence $5\rightarrow4 \rightarrow 3.3 \rightarrow 2 \rightarrow 1$ can still make a reasonable chain with a gradually decreasing trend. The bi-directional constraint is simply too tight to make the chain valid.

\begin{figure}
    \centering
    \vspace{-4mm}
    \includegraphics [width=80mm,height=5cm,
  keepaspectratio]{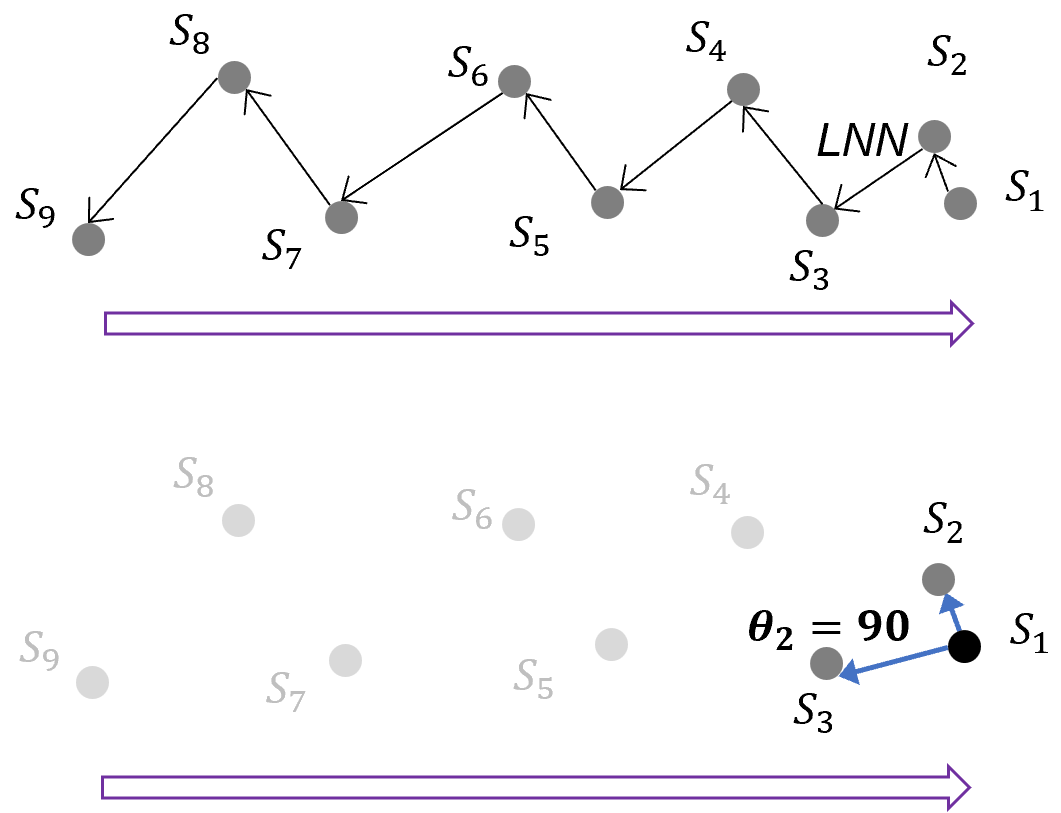}
    \caption{(top) A chain that evolves in a zig-zag pattern due to noise in the data. (bottom) TSC20 fails to find the chain as the first angle is too large.}
    \vspace{-3mm}
    \label{fig:TSC20limitation}
\end{figure}

\subsection{The weakness of TSC20}
The geometric chain definition used in TSC20 \cite{imamura2020matrix} removes the bi-directionl constraint of TSC17 and uses only the backward chain. While this allows the chains to grow longer, as shown in Fig.~\ref{fig:TSC17definition}, unrelated noise patterns can be included in the chain. To solve this, TSC20 added an angle constraint on top of the backward chain. However, we found that the angle constraint can bring in new problems, causing TSC20 to miss obvious chains in some situations, while in some others attaching unwanted patterns to the discovered chain.

\subsubsection{Failure to detect obvious chains} Consider the 2-dimensional example in Fig. \ref{fig:TSC20limitation}.top. Assume that the subsequences show up in the following order in time:
\begin{center}
$S_9$, $S_8$, $S_7$, $S_6$, $S_5$, $S_4$, $S_3$, $S_2$, $S_1$
\end{center}

Although the subsequences evolve in a zig-zag pattern, we can clearly see the trend: they are moving slowly from left to right. However, TSC20 fails to detect the chain.

Following the TSC20 definition, in Fig. \ref{fig:TSC20limitation}.top we first find the backward chain starting from $S_1$ (the arrows point every subsequence to their $LNN$ in the 2-dimensional space), then we check the angles formed between consecutive subsequences based on the anchor subsequence $S_1$. As shown in Fig. \ref{fig:TSC20limitation}.bottom, the first angle $\theta_1=90$ degrees, which is much larger than the suggested threshold 40 degrees in \cite{imamura2020matrix}, and the chain breaks instantly. We can check on all the sub-backward-chains by resetting the anchor to the next subsequence, but unfortunately no sub-chain meets the constraint.

Note that Fig. \ref{fig:TSC20limitation} only shows how fluctuations can fail TSC20 in the 2-dimensional space. In fact, as we will demonstrate later in Section VI, such an evolving pattern is very common in the high-dimensional space, as the noise/fluctuations can drift the subsequences in different directions.

\subsubsection{Attaching unwanted patterns to the chain} 

\begin{figure}
    \centering
    \includegraphics [width=80mm,height=4cm,
  keepaspectratio]{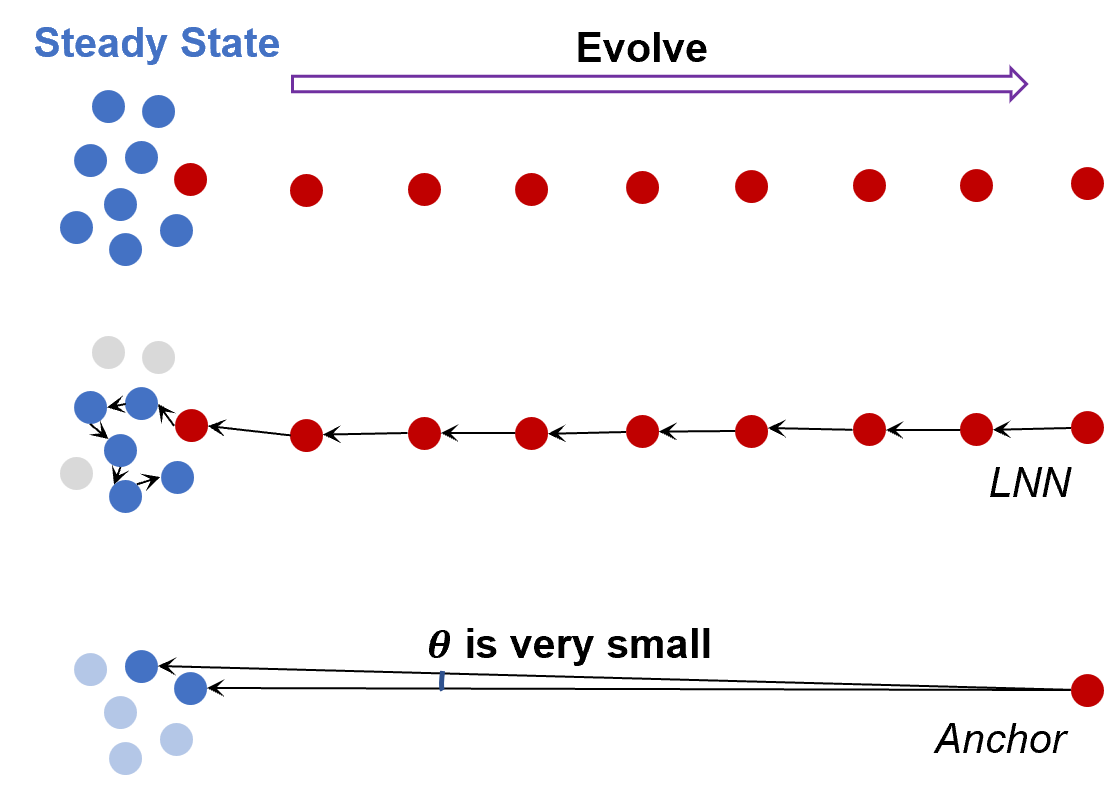}
    \caption{(top) The subsequences diverge from a steady state. (middle) The backward chain includes both the evolving/red subsequences and the steady/blue subsequences. (bottom) TSC20 cannot remove the nodes in the steady state from the evolving chain as their corresponding angles are very small.}
    \vspace{-3mm}
    \label{fig:TSC20limitation3}
\end{figure}

Another problem of TSC20 is that it can sometimes attach unrelated patterns to the chain, deteriorating the chain's quality. Fig. \ref{fig:TSC20limitation3}.top shows an example of this. Here the subsequences are first in a steady state (shown in blue), then they start to drift all the way to the right (shown in red). Ideally, we want the chain discovery algorithm to find all the evolving (red) subsequences, but not those in the steady state (blue), so that the end node of the chain (the first red node on the left) can tell us exactly \textit{when} the system starts to drift (i.e., the change point). This is critical in the prognostics use case, as the time information can help us identify the root cause of the drift. However, as we trace the backward chain shown in Fig. \ref{fig:TSC20limitation3}.middle, we find that all the consecutive nodes along this chain meet the angle constraint of TSC20. The red nodes are evolving in a quasi-linear trend, so the direction angle is close to zero. The angles corresponding to the consecutive blue nodes (shown in Fig. \ref{fig:TSC20limitation3}.bottom) are also very small, as these nodes are very close to each other while far away from the anchor. As a result, the angle constraint of TSC20 cannot exclude the blue nodes from the chain, failing to detect the change point.

We defer further discussions on how this problem affects the quality of the discovered chain in high-dimensional real-world time series to Section VI.

\section{Our Proposed Method}

To address the aforementioned limitations in existing time series chain definitions, we propose a new time series chain discovery method named TSC22. As noted in \cite{imamura2020matrix}, a chain discovery method should consist of two parts: An algorithm to find \emph{all} the chains in the data, and a quality metric used to rank the chains found. In this section, we will first introduce an important concept called the \emph{Incremental Nearest Neighbors}, on top of which we build our chain definitions, then discuss the details of our chain discovery algorithm, and how we rank the discovered chains.

\subsection{Incremental Nearest Neighbors}

\begin{figure}
    \centering
    \includegraphics [width=80mm,height=5cm,
  keepaspectratio]{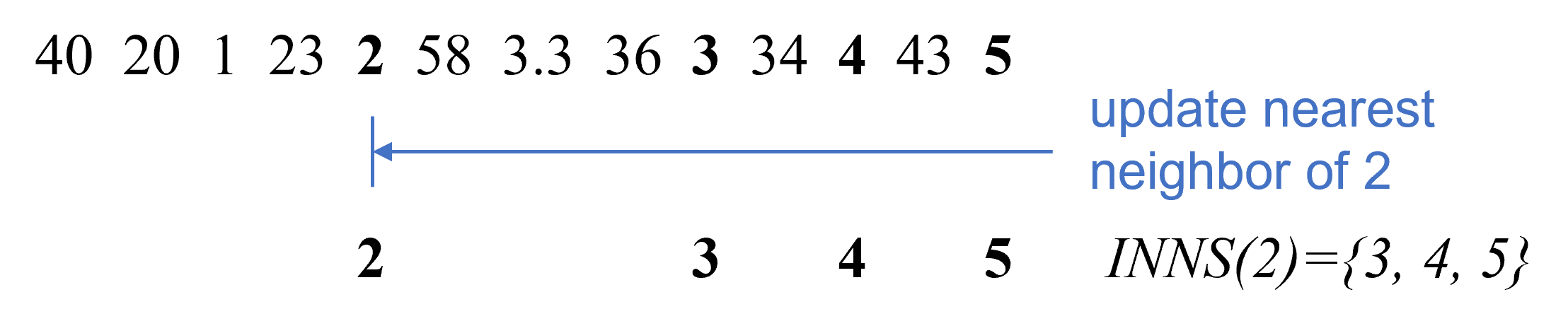}
    \caption{We can obtain an Incremental Nearest Neighbor Set ($INNS$) corresponding to a subsequence by recording its nearest neighbor so far while scanning backwards through the time series.}
    \label{fig:bsf_dp}
\end{figure}

Our chain discovery method leverages an idea from \cite{zhu2020index} to trace how the nearest neighbor of a time series subsequence changes over time. We use the running 1-dimensional example in the previous sections to show how this works in Fig. \ref{fig:bsf_dp}. As we scan from the right all the way to subsequence 2, we store its nearest neighbor subsequence so far (or, its \emph{Incremental Nearest Neighbor}) based on absolute value difference: 5, 4, 3. We call the set $\{3, 4, 5\}$ an \emph{Incremental Nearest Neighbor Set ($INNS$)}, as this set shows how the location of the nearest neighbor of a subsequence changes incrementally when we trace back from the end of the time series.  
\begin{definition}
The \textbf{Incremental Nearest Neighbor Set (INNS)} of a subsequence $S_i$ is a set of subsequences $S_j$ ($i < j \le n-l+1$): $\{S_j| d(S_i, S_j) < d(S_i, S_k) \forall k: j < k \le n-l+1\}$.
\end{definition}

Note that the right nearest neighbor of a subsequence is always in the $INNS$ of that subsequence.

\subsection{Finding All The Chains with Incremental Nearest Neighbors}

Based on the Incremental Nearest Neighbors, we introduce a key component in our chain definition, the critical nodes:
\vspace{-1.5mm}
\begin{definition}
 A time series subsequences $S_i$ of a time series $T$ is a \textbf{Critical Node (CN)} if $S_i \in INNS(LNN(S_i))$.
\end{definition}
\vspace{-1.5mm}

Note that the critical nodes here can be found based on a more relaxed constraint compared to Definition \ref{def:tsc17} (TSC17): a critical node is not necessarily the $RNN$ of its $LNN$, but just need to belong to the incremental nearest neighbor set of its $LNN$. Since the $RNN$ always belongs to the incremental nearest neighbor set, a node in a TSC17 chain is also a critical node in our definition.

We can find all the critical nodes in a time series $T$; we call this a critical node set and denote it as $CN^T$. With this, we can formally define TSC22, a relaxed-bi-directional time series chain:

\begin{definition}
A \textbf{relaxed-bi-directional time series chain (TSC22)} of a time series $T$ is a finite ordered set of time series subsequences: $TSC =[S_{C_1}, S_{C_2}, S_{C_3}, \ldots, S_{C_m}]$ ($C_1 > C_2 >\ldots > C_m$), such that:
\vspace{-1mm}
\begin{enumerate}
    \item for any $1 \leqslant i < m$, we have $LNN(S_{C_{i}}) = S_{C_i+1}$, and
    \item $S_{C_1} \in CN^T$, and
    \item for any $1 < i \leqslant m$, we have $S_{C_j} \in INNS(i)$, where $j = \underset{i < j \le m}{\arg\min}{S_{C_j} \in CN^T}$.
\end{enumerate}
\label{def:tsc22}
\end{definition}
\vspace{-1mm}
From condition (1), we can see that similar to TSC17 and TSC20, our new chain definition TSC22 is also built on top of a backward chain. We call it a ``relaxed" bi-directional chain as our restrictions on the forward direction of the chain is not as strict as that of TSC17: they rely on the incremental nearest neighbor set instead of a single right nearest neighbor.

\begin{figure}
\vspace{-2mm}
    \centering
    \includegraphics [width=80mm,height=5cm,
  keepaspectratio]{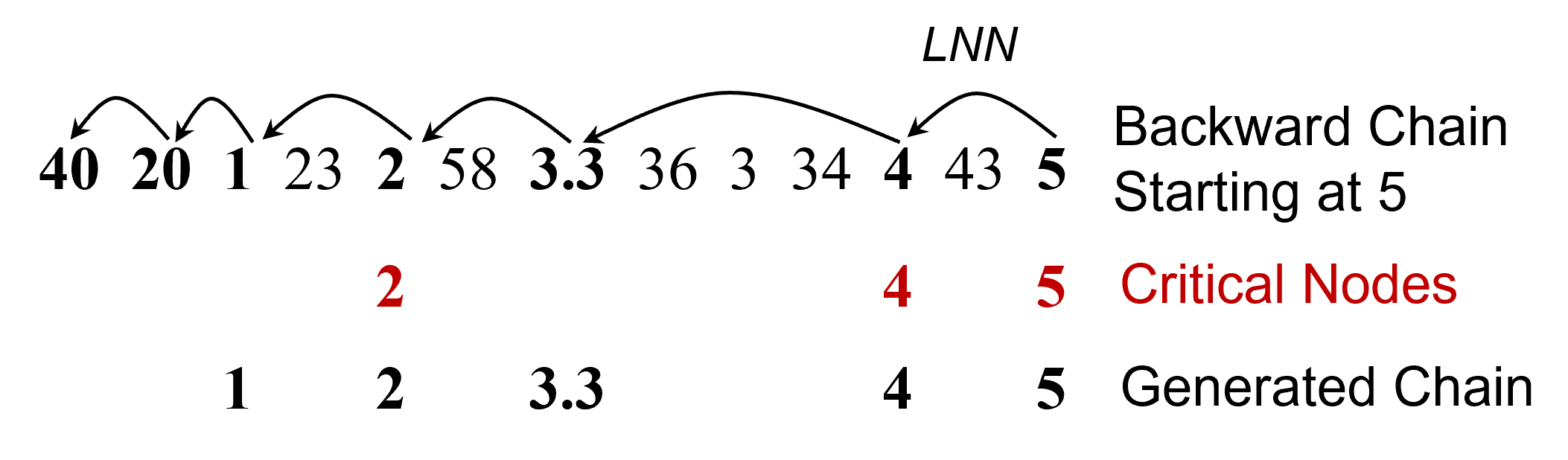}
    \caption{Critical nodes corresponding to the backward chain starting at 5, and the generated chain based on TSC22.}
    \label{fig:criticalnode}
    \vspace{-2mm}
\end{figure}

Fig. \ref{fig:criticalnode} shows a one-dimensional time series as an example. Based on (1) and (2), we can see that TSC22 is a backward chain which starts from a critical node. Let us try to develop a chain from node ``5". As shown in Fig. \ref{fig:criticalnode}, we can see that $LNN(5)=4$ and $5 \in INNS(4)=\{5\}$. So 5 is a critical node, and it is a valid starting node of TSC22. Now we move on to check the next nodes in the backward chain: $LNN(4)=3.3$ and $4 \in INNS(3.3)=\{3, 4, 5\}$, so 4 is a critical node; $LNN(3.3)=2$ but $3.3 \not\in INNS(2)=\{3, 4, 5\}$, so 3.3 is not a critical node, etc. 
Continuing with the process, we can find that the critical nodes on this backward chain are 5, 4 and 2.

Condition (3) requires that for every node $S_{C_i}$ in the chain (except for the starting node), the closest critical node in the chain that appears after $S_{C_i}$ must be an element in $INNS(S_{C_i})$. By checking on every node in the backward chain $5\rightarrow 4\rightarrow 3.3\rightarrow 2\rightarrow1\rightarrow20\rightarrow 40$, we can see that subsequences 5, 4, 3.3, 2, and 1 do meet this condition, but subsequence 20 does not, as $1 \not\in INNC(20) = \{23, 34, 5\}$, so the chain breaks at this location. As such, the chain generated is $5\rightarrow 4\rightarrow 3.3\rightarrow 2\rightarrow 1$. Essentially, condition (3) ensures the ``directionality" of the generated chain. As shown in Fig. \ref{fig:criticalnode}, it makes sure that subsequence 2 is closer to subsequence 1 compared to all the subsequences on the right side of 2, in both the original time series and the generated chain; and 4 is closer to 2 and 3.3 compared to all the subsequences on the right side of subsequence 4, in both the original time series and the generated chain.

By comparing Fig. \ref{fig:criticalnode} with Fig. \ref{fig:TSC17limitation}, we can see that TSC22 allows us to extract a meaningful chain, even when part of the patterns in the chain are flipped due to noise. A diligent reader may notice that our definition is also able to discover the zig-zag chain in Fig. \ref{fig:TSC20limitation}. This shows that TSC22 is more robust than both TSC17 and TSC20 in the face of noisy data. We will demonstrate this claim with more extensive experimental results in Section VI.

The detailed algorithm to find all the chains in a time series based on our TSC22 definition can be found in Algorithm 1. We first compute the $LNN$ and $INNS$ corresponding to all the subsequences by leveraging the STUMP algorithm~\cite{zhu2020index}, which takes essentially the same time as computing a Matrix Profile~\cite{yeh2016matrix}. Then we find all the critical nodes in the time series, and discover chains starting from each critical node in reverse time order. We use a vector to mark whether a critical node has been visited before, so that we can avoid repeated computations. Once we finish the exploration, we store all the sub-chains corresponding to each discovered chain in the our results. The time and space complexity of our algorithm is identical to that of TSC20~\cite{imamura2020matrix}, and the time used to compute the chains given the Matrix Profile is negligible compared to the time required to compute the Matrix Profile. 

\begin{algorithm}[t]
    \caption{Find All TSC22 Chains in A Time Series}
 \begin{algorithmic}[1]
    \STATE \textbf{Input}: Time Series $T$
    \STATE \textbf{Output}: All Chain Set $C_{TSC22}$
    \\{\color{blue}/*Compute $LNN$ and $INNS$ with STUMP\cite{zhu2020index}}*/
    \STATE $LNN, INNS = $ComputeLNNandINNS($T$)
    \\{\color{blue}/* Find All Critical Nodes in $T$ */}
    \STATE $S_{critical}= $FindAllCriticalNode($LNN$, $INNS$)
    \\{\color{blue}/*Vector to mark whether an index has been visited.*/}
    \STATE VISITED.fill(len($LNN$), False);
    \\{\color{blue}/*Visit critical points in reverse time order.*/}
    \FOR{$S$ in $S_{critical}$.reverse()}
     \\{\color{blue}/*if $S$ is visited before, then the chain grown from it must be the sub-chain of a chain that we have already discovered. Skip.*/}
    \IF{VISITED($S$)}
      \STATE \textbf{continue}
    \ENDIF
    \STATE Chain=[$S$], $S'$=$S$, VISITED[Index[S]]=True;
    \WHILE{LNN($S'$) is not null}
      \\{\color{blue}/* check if $S$ is in INNS(LNN[$S'$])/}
      \IF{$S$ not in LNN[$S'$]}
        \STATE \textbf{break}
      \ELSE 
        \STATE Chain.add(LNN[$S'$]), S'=LNN[$S'$];
        \STATE VISITED[Index[$S'$]]=True;
        \IF{$S'$ in $S_{critical}$}
           S=$S'$;
        \ENDIF
      \ENDIF 
    \ENDWHILE
     \\{\color{blue}/* update all sub-chains of Chain to their corresponding index in the all-chain set  */}
    \STATE $C_{TSC22}$ = $C_{TSC22}$.update(AllSubChains(Chain))
    \ENDFOR
    
    \STATE \textbf{return} $C_{TSC22}$
    
  \end{algorithmic}
\end{algorithm}


 \vspace{-1mm}
\subsection{Ranking}
Now that we have all the TSC22 chains discovered, we need a mechanism to measure their quality, so that we can effectively rank them. TSC17 ranks the discovered chains by their length, which is not always effective, especially when a large amount of noise is present in the time series. As noted in \cite{imamura2020matrix}, sometimes TSC17 can discover a long chain even in steady data which does not display any evolving trend, so there is no way we can tell whether there is any meaningful chain in the data merely from the length of the top chain. Ideally, a high quality chain should have the following properties:
\begin{itemize}
    \item \textbf{High Divergence}  The first and the last nodes in the chain should be sufficiently dissimilar.
    \item \textbf{Gradual Change}  The consecutive nodes in the chain should be very similar to each other.
    \item \textbf{Purity} The chain should not include unrelated patterns.
    \item \textbf{Long Length}  We would like the chain to be long, so that it has a good coverage of the drifting patterns.
\end{itemize} 
Note that these quality perspectives do not always agree with each other. One chain can evolve very quickly, leading to high divergence but low graduality and short length, while another long chain can consist of patterns that are almost identical to each other but do not show any evolving trend. To resolve this, we develop two quality metrics that take all the properties into account, and design a two-stage ranking algorithm based on these quality metrics to make sure that a high quality chain can stand out. 

\textbf{Effective Length} Inspired by TSC20~\cite{imamura2020matrix}, we compute an ``effective length" metric $L_{eff}$, which measures both divergence and graduality at the same time:
\begin{align}
    L_{eff} = \lfloor d(S_{C_1}, S_{C_m})/ \underset{1 \leq i \leq m-1}\max d(S_{C_i},S_{C_{i+1}})\rceil,
\end{align} 
\noindent where $\lfloor.\rceil$ denotes rounding to the nearest integer. The numerator is the distance between the first node and the last node in the chain, and the denominator is the maximum distance between all pairs of consecutive nodes in the chain. One can imagine that if a chain evolves in a linear trace with a uniform pace, $L_{eff}$ will become the length of the chain. The metric essentially tells us the approximate number of steps we need to take to reach the end node from the start node if we are moving in a linear trace, with a distance per step that is equal to the maximum consecutive distances between the nodes in the chain. A chain with high divergence as well as good graduality should show a high $L_{eff}$ value; if the chain includes noise, we should have $L_{eff}$ $\approx 1$.

As $L_{eff}$ is an integer, there will be ties. We first rank the chains by their $L_{eff}$ scores, then for those with the highest $L_{eff}$ score, we do a fine-grain ranking based on the squared sum of correlations.

\textbf{Correlation Length} We call our metric the Correlation Length, which is computed as follows:
\vspace{-1mm}
\begin{equation}
    L_{corr} = \sum_{i=1}^{m-1} |Corr(S_{C_i}, S_{C_{i+1}})|Corr(S_{C_i}, S_{C_{i+1}}).
    \vspace{-1mm}
\end{equation}
\noindent where $Corr(.)$ is the Pearson Correlation Coefficient of the z-normalized subsequences, and the values are in range $[-1, 1]$. When two subsequences are very similar to each other, their Pearson Correlation Coefficient is close to 1; when the pair includes noise, the Pearson Correlation Coefficient is normally smaller than 0.5. Here we multiply the Pearson Correlation Coefficient with its absolute value just to enlarge the difference between similar pairs and dissimilar pairs. We call this metric the correlation ``length"  because if consecutive nodes in a chain are very similar to each other, $L_{corr}$ will be very close to the actual length of the chain. As a result, longer chains with highly similar subsequences will be preferred at the fine-grain ranking stage.
Compare to the ranking method of TSC17, which only considers the length of a chain, and that of TSC20 which considers only the divergence and graduality, our ranking method has a better coverage over all quality perspectives of a chain. As we will demonstrate in the next section, the new ranking approach, together with the more robust chain definition, allows TSC22 to effectively find meaningful chains in a variety of real-world and synthetic datasets, even when a large amount of noise is present in the data.
 \vspace{-1mm}
\section{EMPIRICAL EVALUATION}\label{sec6}
In this section, we demonstrate that the proposed method is more robust than the current state-of-the-art time series chain discovery methods, \textbf{TSC17 (ICDM'17)}\cite{zhu2017matrix} and \textbf{TSC20 (KDD'20)}\cite{imamura2020matrix} on both real-world and synthetic data. To ensure fair comparison, we use the original source code of both TSC17 and TSC20 in Matlab, and we use the default direction angle for TSC20.  To make the results easily reproducible, we built a supporting webpage \cite{supportingwebpage22} that contains all the data and code used in this section.


We will first analyze the performance of the three chain discovery methods \textit{qualitatively} with several case studies on \textbf{real-world} data, then we will do a \textit{quantitative} comparison on a large-scale \textbf{synthetic} dataset, where we can compute the performance metrics based on absolute ground truth.

To begin with, let us consider a time series that records a penguin's diving activity.

\subsection{Case Study: Robustness Analysis on Penguin Activity Data}
\vspace{-1.5mm} 
\subsubsection{Clean Data}

\begin{figure}[h]
    \centering
    \vspace{-2mm} 
    \includegraphics [width=80mm]{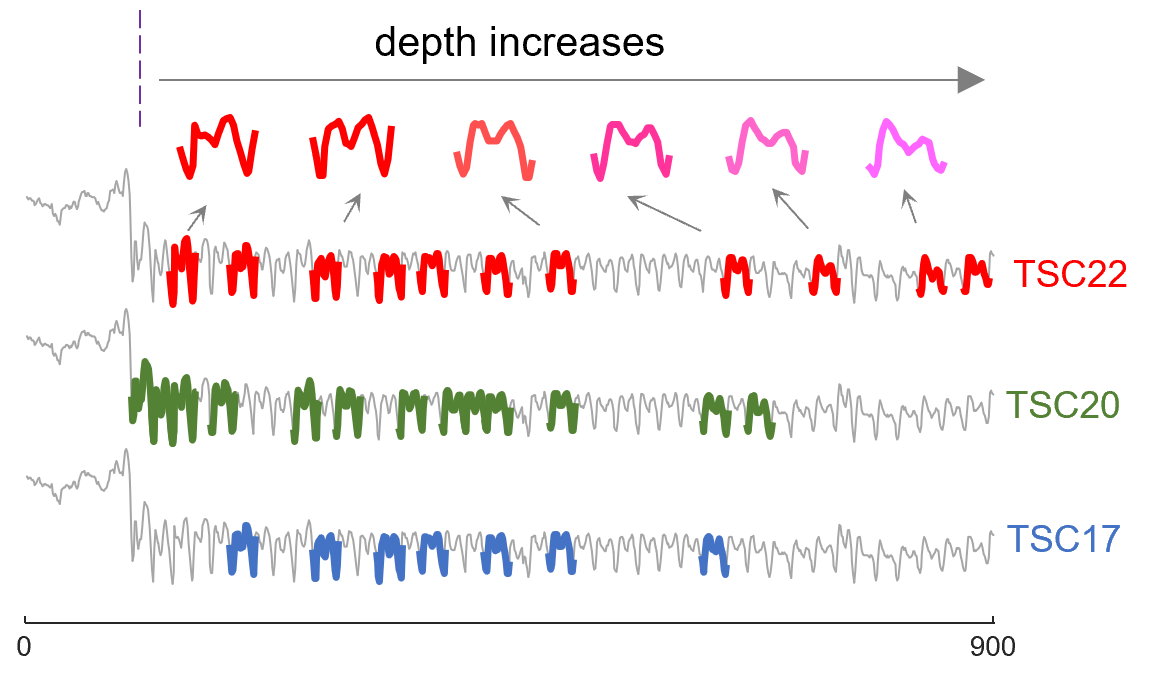}
    \caption{Chains detected in a clean penguin activity time series as the penguin dives into the water. TSC22 is able to find a high-quality chain that covers the whole diving range; TSC20 misses a few nodes at the end and TSC17 misses nodes on both sides.}
    \vspace{-2mm} 
    \label{fig:penguin_clean}
\end{figure}

Here we use the penguin activity time series in \cite{zhu2017matrix}, which shows the x-axis acceleration of the bird as it moves. The subsequence length is 25. Fig. \ref{fig:penguin_clean} shows a 22.5s snippet of the data (recorded at 40Hz) as the penguin dives into the water. The bird reaches the water at around 2 seconds (the 80th data point), and then the depth value starts to increase. As shown in Fig. \ref{fig:penguin_clean}, TSC22 finds a high-quality chain that clearly shows an evolving trend: at the beginning the first peak in the pattern is slightly lower than the second peak, then over time the first peak grows higher and the second peak becomes weaker. This chain indicates how the penguin adjusts its flapping to balance between the buoyancy and the increasing water pressure \cite{zhu2017matrix}. The chain found by TSC22 covers the full diving range, while the TSC20 chain misses a few nodes at the end and the TSC17 chain misses a few more on both sides. Though the chains found by TSC20 and TSC17 can still show the evolving trend of the patterns, they fail to indicate the start/end time of the evolving procedure.

Since we are reporting the top-1 chain found by all three methods, the reader may wonder whether it is the constraints of TSC17 and TSC20 that prevents them from finding the TSC22 chain in Fig. \ref{fig:penguin_clean}, or it is their ranking mechanisms that prefer other chains over the TSC22 chain. By close inspection, we can see that the TSC22 chain does not satisfy the definition of either TSC17 or TSC20. Similar as in Fig. \ref{fig:TSC17limitation}, most of the consecutive nodes on the chain do not satisfy the bi-directional nearest neighbor constraint of TSC17 because of the natural fluctuations in the data (e.g., the penguin may have changed its moving direction to avoid other animals). And as we trace the TSC22 chain backwards, we find that the directional angle formed by the first three nodes on this chain (recall Fig. \ref{fig:TSC20limitation}) is 114 degrees, which is much higher than the angle threshold of TSC20 (40 degrees), so the chain breaks instantly. This example verifies the weaknesses of TSC17 and TSC20 in high-dimensional, real-world data with natural fluctuations and demonstrates the robustness of TSC22 in such data.

\begin{figure}[ht]
    \centering
     \vspace{-4mm}
    \includegraphics [width=80mm]{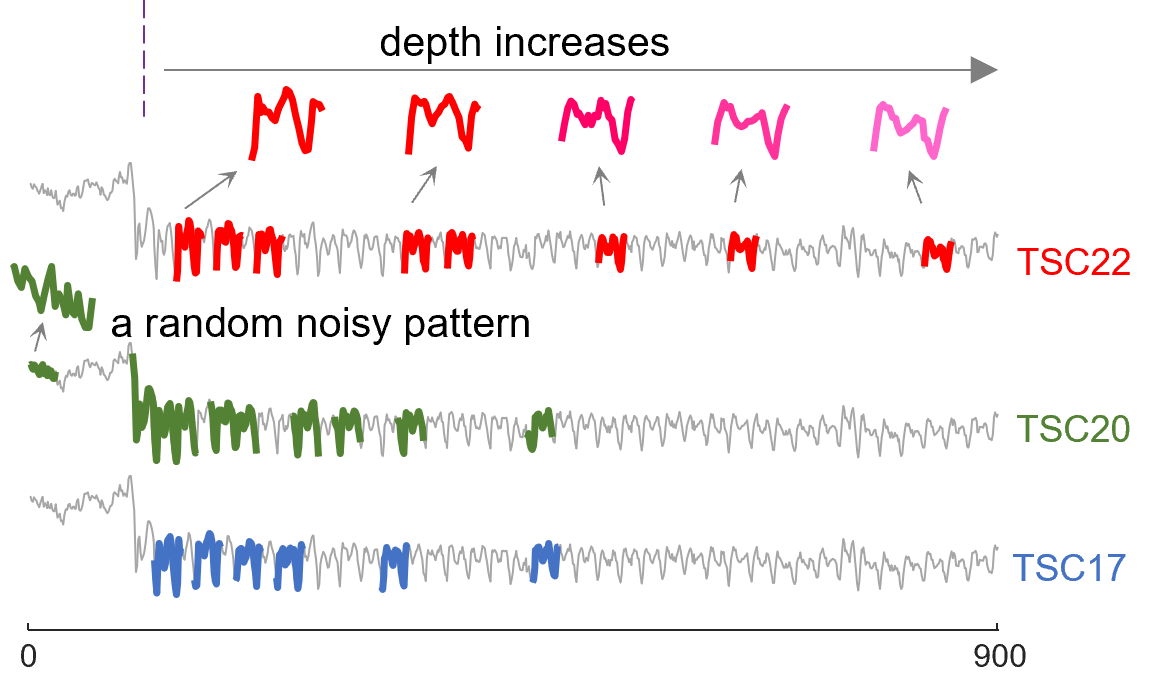}
    \caption{Chains detected in the penguin activity time series with a small amount of injected background noise. The patterns still show a very clear evolving trend. TSC22 is able to find a chain that covers the whole diving period, while those found by TSC20 and TSC17 are much shorter. A random noisy subsequence is included in the TSC20 chain, which is very undesirable.}
     \vspace{-3mm}
    \label{fig:penguin_noisy}
\end{figure}

\subsubsection{Data with Random Background Noise}

To further stress-test the robustness of of the three methods in concern, in Fig. \ref{fig:penguin_noisy} we added a small amount ($\pm 0.08$ magnitude) of random noise to the penguin data, which simulates the additional background noise from the sensor. We can see that TSC22 is still able to find a high-quality chain that ranges across the whole diving procedure in this case, and the patterns discovered still show a very clear evolving trend even with the additional noise. However, TSC20 and TSC17 both fail to capture this chain, and their top chains can now only cover half of the evolving range. Further note that TSC20 attaches a random noisy pattern to the top chain: this indicates the weakened filtering power of its direction angle constraint as the backward chain grows longer.  
\vspace{-1mm}
\vspace{-1mm}
\subsection{Case Study: Change Point Detection with Tilt Table Data}

\begin{figure}[h]
    \centering
    \vspace{-3mm}
    \includegraphics [width=85mm,height=5cm,
  keepaspectratio]{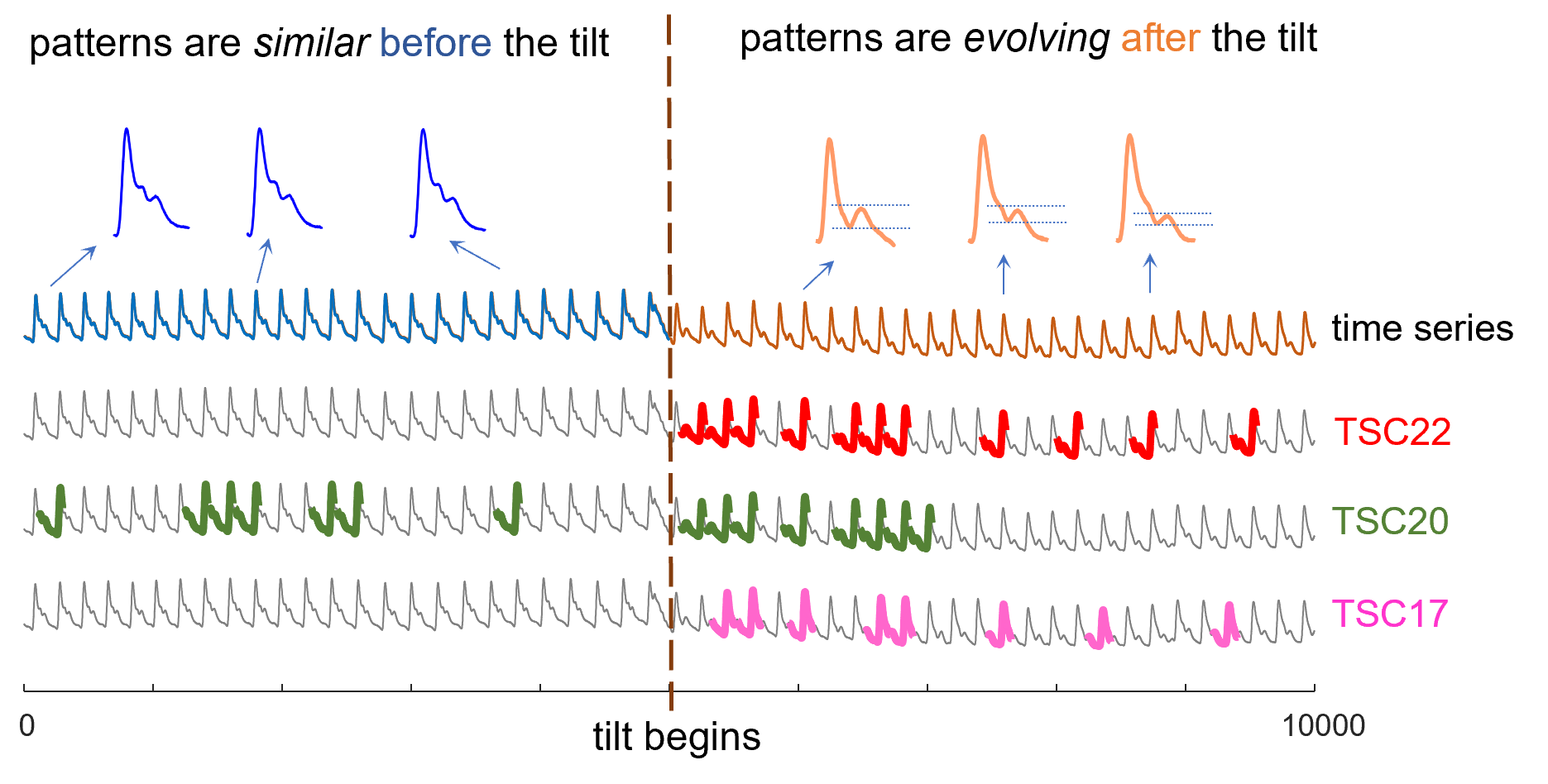}
    \caption{The ABP time series of a patient in a tilt table experiment. The table is first flat, then it starts to tilt. The top chains detected by TSC22 and TSC17 only include the evolving patterns after the tilt, while TSC20 detects a chain that spans across the boundary, failing to locate the change point.}
    \label{fig:tm_se}
\end{figure}

In this section, we use the tilt table data in \cite{zhu2017matrix} to compare how the three methods in consideration perform when a system transitions from a steady status to a drifting status. Fig. \ref{fig:tm_se} shows the the arterial blood pressure (ABP) signal of a patient lying on a tilt table. Here we use a subsequence length of 180, roughly the length of a cardiac cycle. The table is flat at the beginning, and then starts to tilt. We would expect chains to be detected \emph{after} the the tilt and no chain should appear \emph{before} the tilt. We also expect a detected chain not to cross the tilt/non-tilt boundary, as the repeated patterns before the tilt would deteriorate the purity and interpretability of the chain.

Note that this kind of data is very typical in many other domains, especially in prognostics, where a system operates at a stable status at the beginning, and then start to deteriorate. It is essential to correctely locate the change point (i.e., the beginning of the drift), so people can use that information to identify implicit mechanisms that could have led to the drift, preventing systematic failures at an early stage.

In Fig. \ref{fig:tm_se}, we can see that both TSC22 and TSC17 successfully discover chains that cover only the right section of the data after the tilt, while TSC20 detected a chain that spans across the tilting section and the steady section. It is not hard to understand why TSC20 fails to produce a pure chain in this case (recall Fig. \ref{fig:TSC20limitation3}): the direction angle formed between any pair of subsequences before the tilt based on the anchor point would be close to zero, as these subsequences are almost identical. The single-directional nature of TSC20 simply cannot stop the backward chain from growing when it reaches the stable section of the data. In contrast, the bi-directional restrictions in both TSC22 and TSC17 effectively prevent their chains from growing into the steady section. The example further demonstrates the usefulness of TSC22: it can precisely locate \textit{when} the system starts to drift, allowing analyzers to easily find the implicit cause of the drift by investigating the events happening around that time.

\subsection{Case Study: Finding Chains across Multiple Exercise Stages in Human Gait Force Data}

\begin{figure}[h]
    \centering
    \includegraphics [width=85mm,height=5cm,
  keepaspectratio]{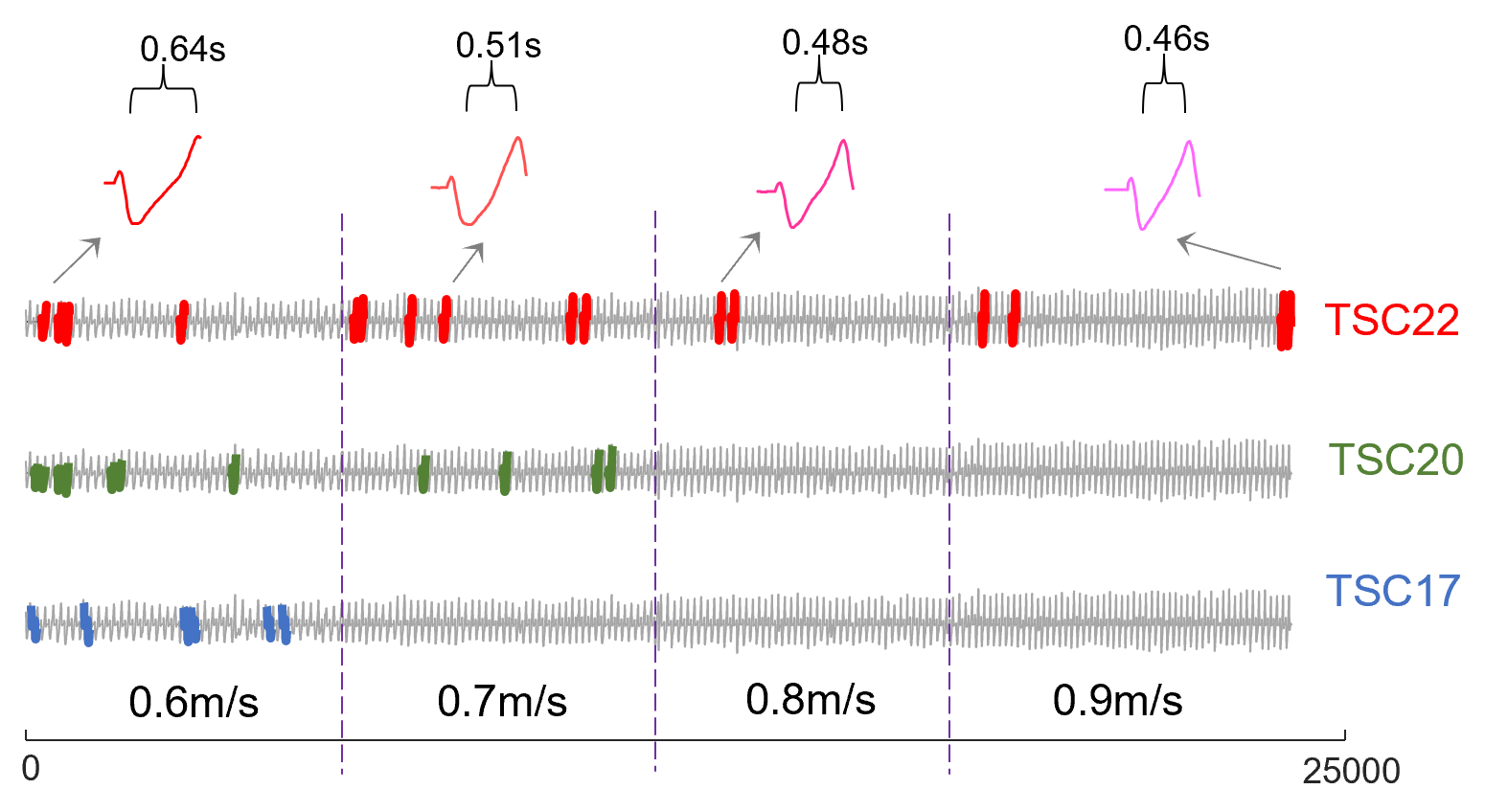}
    \caption{A human gait force time series collcted on a threadmill with increasing speeds. The chain detected by TSC22 can capture evolving patterns across all four different speed zones, while TSC20 can only cover two and TSC17 only one.}
    \label{fig:tm}
    \vspace{-5mm}
\end{figure}

We next evaluate the chain discovery methods on a human gait force time series. As shown in Fig. \ref{fig:tm}, the time series represents the posterior-anterior direction force detected by the sensor on a split-belt treadmill, which operates at four different running speeds: 0.6, 0.7, 0.8, and 0.9 m/s. Here we use a subsequence length of 100 as it is close to a cycle in the data. We can see from Fig. \ref{fig:tm} that the chain detected by TSC22 spans over all different speed zones. The sharpening dips in the chain patters indicate that as the speed increases, the force changes more quickly, and the participant's foot spend less time on the threadmill floor. The shortening period between the dip and the peak also indicate that the running paces become faster. However, despite the fact that the patterns are evolving across the whole time series, TSC20 and TSC17 can only detect chains on a small portion of the data, failing to cover the complete range.

Why do TSC20 and TSC17 fail in this case? Note that as the speed is increasing at a piecewise-constant fashion, the patterns within the same speed zone are relatively similar to each other, evolving at a much smaller pace. As the participant's movements naturally include fluctuations, a lot of patterns get flipped (recall Fig. \ref{fig:TSC17limitation}) within a speed zone, and the overall evolving trace is similar to the zig-zag chains shown in Fig. \ref{fig:introexample}.c and Fig. \ref{fig:TSC20limitation} instead of the one in Fig. \ref{fig:introexample}.a. Note that the effect of the fluctuations is even more severe now as we are facing high-dimensional data. As a sanity check, we find that none of the nodes on the TSC22 chain in Fig. \ref{fig:tm} meet the bi-directional nearest neighbor constraint of TSC20, and the direction angle formed by the last three nodes on the chain (i.e., the first three nodes in the backward chain) is 86 degrees, much larger than the angle limit of 40 degrees in TSC20. This example further demonstrates the superior robustness of TSC22 on real-world time series data that naturally come with many fluctuations and noise.
\vspace{-2mm}
\subsection{Quantitative Robustness Analysis} 
In the previous sections, we have demonstrated \textit{qualitatively} that TSC22 outperforms TSC20 and TSC17 in various real-world scenarios. However, the lack of absolute ground truth labels (i.e., the exact location of each pattern of the chain in the time series) and the potential ambiguity in these labels~\cite{wu2021current} make it hard to perform any reasonable \text{quantitative} analysis based on real-world data. In order to compute quantifiable performance metrics such as precision, recall, etc., we created a large scale synthetic dataset in which we manually embed the chain patterns, and test whether the three methods in concern can accurately locate these patterns. In this section, we will briefly describe the dataset, formally define the performance metrics, and compare the three methods in concern both with and without the ranking mechanism. 

\begin{figure}
    \centering
    \vspace{-2mm} 
    \includegraphics [width=85mm,height=5.5cm,
  keepaspectratio]{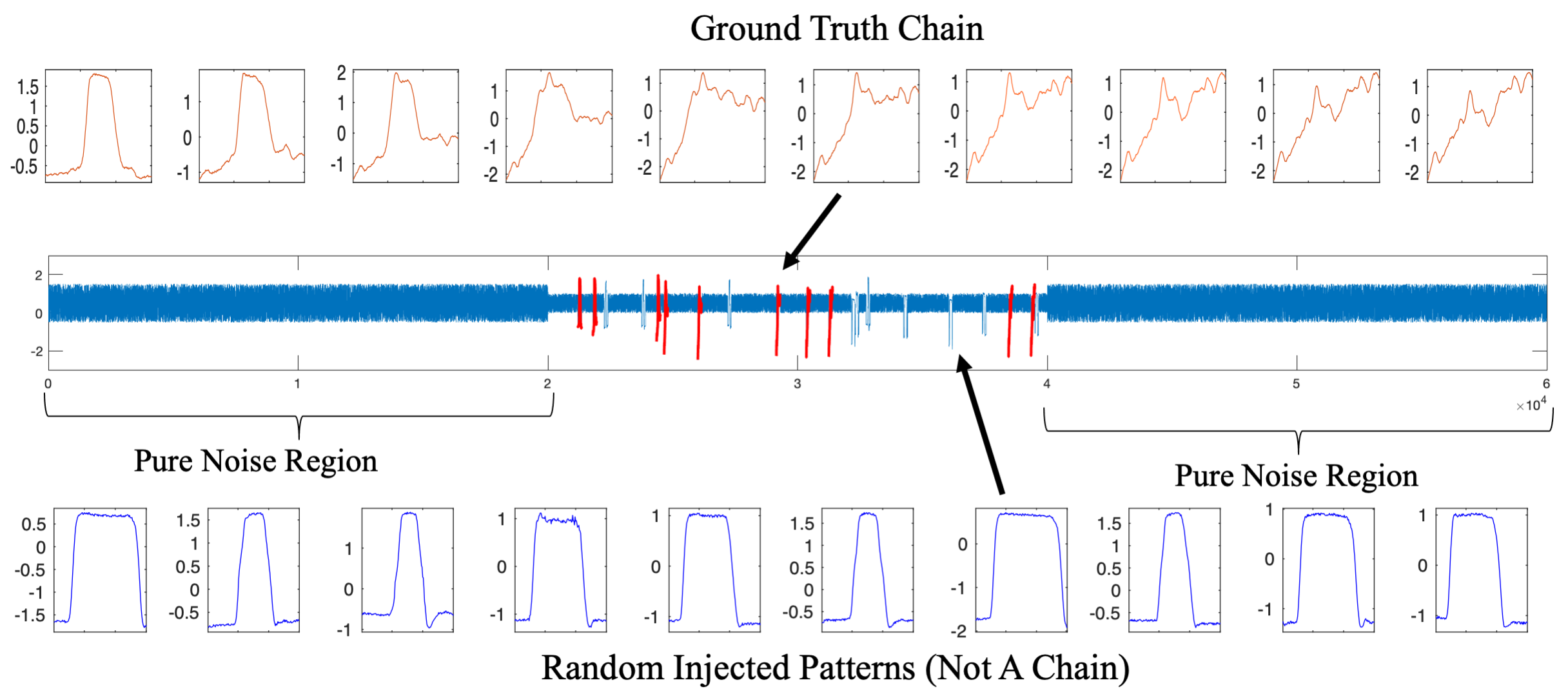}
  \vspace{-2mm} 
    \caption{An Example of our Generated Time Series}
    \label{fig:syn_example}
    \vspace{-5mm}
\end{figure}

Fig.~\ref{fig:syn_example} shows an example time series we created. Here the first node in the ground truth chain (shown in Fig.~\ref{fig:syn_example}.top) is randomly sampled from a dataset in the the UCR time series classification archive \footnote{\href{https://www.timeseriesclassification.com/dataset.php}{https://www.timeseriesclassification.com/dataset.php}}, and the last node is a random-walk pattern. The chain consists of 10 nodes, evolving linearly from the first node to the last. We embeded the chain nodes into a random-noise time series at randomly locations while keeping their order. To increase the difficulty of the task, we added random noise on top of the data to distort the patterns, and embedded 10 additional patterns that are similar to the first node (sampled from the same dataset) but irrelevant to the evolving trend (shown in Fig.~\ref{fig:syn_example}.bottom) into the time series. Two random noise sections are added to the beginning and the end of the generated time series to simulate the real-world scenario where chains do not span across the whole time series. We used seventeen datasets in the the UCR time series archive to construct the synthetic time series, covering all the sensor, ECG and simulated shape data with instance length ranging from 50 to 500, and omitted the datasets containing high-frequency patterns from which the generated chains would be harder to interpret visually. For each dataset we generated five different time series and measured the average performance of each chain discovery method on these time series.

Inspired by \cite{zhu2017matrix}, we use the F1-score to measure the quality of the detected chains. We define a \textbf{hit} if a node (i.e., subsequence) in the detected chain overlaps with the ground truth by more than 50\%. We compute the precision, recall and the F1-score as follows: $$\text{recall} = \frac{\text{number of hits}}{\text{length of the ground truth chain}},$$
$$\text{precision} = \frac{\text{number of hits}}{\text{length of the detected chain}},$$ 

$$\text{F1-score} = \frac{2\text{precision}\cdot\text{recall}}{\text{precision}+\text{recall}}.$$ 

As all TSC methods consist of two steps: chain detection and ranking, we ran experiments to compare the performance of these methods both with and without the ranking step.

\begin{table}[h]
\vspace{-2mm}

\caption{F1-Score Performance vs. Baselines (with Ranking)}
\vspace{-2mm}
\scalebox{0.55}{
\begin{tabular}{llllllllll}
\cline{2-10}
\multicolumn{1}{l|}{}                       & \multicolumn{3}{c|}{TSC17}                                                                   & \multicolumn{3}{c|}{TSC20}                                                                         & \multicolumn{3}{c|}{Ours (TSC22)}                                                                                       \\ \hline
\multicolumn{1}{|l|}{Pattern Source}        & \multicolumn{1}{c|}{recall} & \multicolumn{1}{c|}{precision} & \multicolumn{1}{c|}{F1-score} & \multicolumn{1}{c|}{recall} & \multicolumn{1}{c|}{precision} & \multicolumn{1}{c|}{F1-score}       & \multicolumn{1}{c|}{recall}         & \multicolumn{1}{c|}{precision}      & \multicolumn{1}{c|}{F1-score}       \\ \hline
\multicolumn{1}{|l|}{Cl. Con.} & \multicolumn{1}{l|}{0.220}  & \multicolumn{1}{l|}{0.258}     & \multicolumn{1}{l|}{0.237}    & \multicolumn{1}{l|}{0.520}  & \multicolumn{1}{l|}{0.865}     & \multicolumn{1}{l|}{0.638}          & \multicolumn{1}{l|}{0.580}          & \multicolumn{1}{l|}{0.730}          & \multicolumn{1}{l|}{\textbf{0.641}} \\ \hline
\multicolumn{1}{|l|}{BME}                   & \multicolumn{1}{l|}{0.000}  & \multicolumn{1}{l|}{0.000}     & \multicolumn{1}{l|}{0.000}    & \multicolumn{1}{l|}{0.540}  & \multicolumn{1}{l|}{0.685}     & \multicolumn{1}{l|}{0.602}          & \multicolumn{1}{l|}{0.560}          & \multicolumn{1}{l|}{0.778}          & \multicolumn{1}{l|}{\textbf{0.647}} \\ \hline
\multicolumn{1}{|l|}{TwoPatterns}           & \multicolumn{1}{l|}{0.000}  & \multicolumn{1}{l|}{0.000}     & \multicolumn{1}{l|}{0.000}    & \multicolumn{1}{l|}{0.280}  & \multicolumn{1}{l|}{0.431}     & \multicolumn{1}{l|}{0.335}          & \multicolumn{1}{l|}{0.640}          & \multicolumn{1}{l|}{0.800}          & \multicolumn{1}{l|}{\textbf{0.710}} \\ \hline
\multicolumn{1}{|l|}{ECG5000}               & \multicolumn{1}{l|}{0.340}  & \multicolumn{1}{l|}{0.342}     & \multicolumn{1}{l|}{0.337}    & \multicolumn{1}{l|}{0.800}  & \multicolumn{1}{l|}{0.788}     & \multicolumn{1}{l|}{0.791}          & \multicolumn{1}{l|}{0.820}          & \multicolumn{1}{l|}{0.975}          & \multicolumn{1}{l|}{\textbf{0.889}} \\ \hline
\multicolumn{1}{|l|}{CBF}                   & \multicolumn{1}{l|}{0.160}  & \multicolumn{1}{l|}{0.200}     & \multicolumn{1}{l|}{0.178}    & \multicolumn{1}{l|}{0.260}  & \multicolumn{1}{l|}{0.353}     & \multicolumn{1}{l|}{0.292}          & \multicolumn{1}{l|}{0.780}          & \multicolumn{1}{l|}{0.964}          & \multicolumn{1}{l|}{\textbf{0.856}} \\ \hline
\multicolumn{1}{|l|}{TwoLeadECG}            & \multicolumn{1}{l|}{0.340}  & \multicolumn{1}{l|}{0.378}     & \multicolumn{1}{l|}{0.358}    & \multicolumn{1}{l|}{0.780}  & \multicolumn{1}{l|}{0.919}     & \multicolumn{1}{l|}{\textbf{0.838}} & \multicolumn{1}{l|}{0.660}          & \multicolumn{1}{l|}{0.796}          & \multicolumn{1}{l|}{0.718}          \\ \hline
\multicolumn{1}{|l|}{ECG200}                & \multicolumn{1}{l|}{0.000}  & \multicolumn{1}{l|}{0.000}     & \multicolumn{1}{l|}{0.000}    & \multicolumn{1}{l|}{0.560}  & \multicolumn{1}{l|}{0.825}     & \multicolumn{1}{l|}{\textbf{0.666}} & \multicolumn{1}{l|}{0.520}          & \multicolumn{1}{l|}{0.870}          & \multicolumn{1}{l|}{0.642}          \\ \hline
\multicolumn{1}{|l|}{ECGFiveDays}           & \multicolumn{1}{l|}{0.500}  & \multicolumn{1}{l|}{0.578}     & \multicolumn{1}{l|}{0.536}    & \multicolumn{1}{l|}{0.700}  & \multicolumn{1}{l|}{0.829}     & \multicolumn{1}{l|}{0.748}          & \multicolumn{1}{l|}{0.800}          & \multicolumn{1}{l|}{1.000}          & \multicolumn{1}{l|}{\textbf{0.886}} \\ \hline
\multicolumn{1}{|l|}{FreezerSmallTrain}     & \multicolumn{1}{l|}{0.320}  & \multicolumn{1}{l|}{0.356}     & \multicolumn{1}{l|}{0.337}    & \multicolumn{1}{l|}{0.720}  & \multicolumn{1}{l|}{0.866}     & \multicolumn{1}{l|}{0.779}          & \multicolumn{1}{l|}{0.820}          & \multicolumn{1}{l|}{0.978}          & \multicolumn{1}{l|}{\textbf{0.889}} \\ \hline
\multicolumn{1}{|l|}{RegularSmallTrain}     & \multicolumn{1}{l|}{0.320}  & \multicolumn{1}{l|}{0.292}     & \multicolumn{1}{l|}{0.302}    & \multicolumn{1}{l|}{0.640}  & \multicolumn{1}{l|}{0.840}     & \multicolumn{1}{l|}{0.721}          & \multicolumn{1}{l|}{0.720}          & \multicolumn{1}{l|}{0.980}          & \multicolumn{1}{l|}{\textbf{0.822}} \\ \hline
\multicolumn{1}{|l|}{Trace}                 & \multicolumn{1}{l|}{0.240}  & \multicolumn{1}{l|}{0.292}     & \multicolumn{1}{l|}{0.263}    & \multicolumn{1}{l|}{0.560}  & \multicolumn{1}{l|}{0.905}     & \multicolumn{1}{l|}{0.683}          & \multicolumn{1}{l|}{0.700}          & \multicolumn{1}{l|}{0.920}          & \multicolumn{1}{l|}{\textbf{0.781}} \\ \hline
\multicolumn{1}{|l|}{Wafer}                 & \multicolumn{1}{l|}{0.340}  & \multicolumn{1}{l|}{0.400}     & \multicolumn{1}{l|}{0.367}    & \multicolumn{1}{l|}{0.600}  & \multicolumn{1}{l|}{0.785}     & \multicolumn{1}{l|}{0.660}          & \multicolumn{1}{l|}{0.600}          & \multicolumn{1}{l|}{0.775}          & \multicolumn{1}{l|}{\textbf{0.673}} \\ \hline
\multicolumn{1}{|l|}{Plane}                 & \multicolumn{1}{l|}{0.700}  & \multicolumn{1}{l|}{0.780}     & \multicolumn{1}{l|}{0.736}    & \multicolumn{1}{l|}{0.900}  & \multicolumn{1}{l|}{0.920}     & \multicolumn{1}{l|}{0.909}          & \multicolumn{1}{l|}{0.860}          & \multicolumn{1}{l|}{1.000}          & \multicolumn{1}{l|}{\textbf{0.921}} \\ \hline
\multicolumn{1}{|l|}{SonyAI}                & \multicolumn{1}{l|}{0.000}  & \multicolumn{1}{l|}{0.000}     & \multicolumn{1}{l|}{0.000}    & \multicolumn{1}{l|}{0.360}  & \multicolumn{1}{l|}{0.535}     & \multicolumn{1}{l|}{0.429}          & \multicolumn{1}{l|}{0.700}          & \multicolumn{1}{l|}{0.971}          & \multicolumn{1}{l|}{\textbf{0.810}} \\ \hline
\multicolumn{1}{|l|}{SonyAI2}               & \multicolumn{1}{l|}{0.000}  & \multicolumn{1}{l|}{0.000}     & \multicolumn{1}{l|}{0.000}    & \multicolumn{1}{l|}{0.140}  & \multicolumn{1}{l|}{0.224}     & \multicolumn{1}{l|}{0.170}          & \multicolumn{1}{l|}{0.360}          & \multicolumn{1}{l|}{0.511}          & \multicolumn{1}{l|}{\textbf{0.418}} \\ \hline
\multicolumn{1}{|l|}{Lightning7}            & \multicolumn{1}{l|}{0.000}  & \multicolumn{1}{l|}{0.000}     & \multicolumn{1}{l|}{0.000}    & \multicolumn{1}{l|}{0.300}  & \multicolumn{1}{l|}{0.481}     & \multicolumn{1}{l|}{0.366}          & \multicolumn{1}{l|}{0.480}          & \multicolumn{1}{l|}{0.814}          & \multicolumn{1}{l|}{\textbf{0.592}} \\ \hline
\multicolumn{1}{|l|}{UME}                   & \multicolumn{1}{l|}{0.160}  & \multicolumn{1}{l|}{0.200}     & \multicolumn{1}{l|}{0.178}    & \multicolumn{1}{l|}{0.540}  & \multicolumn{1}{l|}{0.814}     & \multicolumn{1}{l|}{\textbf{0.643}} & \multicolumn{1}{l|}{0.520}          & \multicolumn{1}{l|}{0.752}          & \multicolumn{1}{l|}{0.610} \\ \hline
\multicolumn{1}{|l|}{Average}               & \multicolumn{1}{l|}{0.214}  & \multicolumn{1}{l|}{0.240}     & \multicolumn{1}{l|}{0.225}    & \multicolumn{1}{l|}{0.541}  & \multicolumn{1}{l|}{0.710}     & \multicolumn{1}{l|}{0.604}          & \multicolumn{1}{l|}{\textbf{0.654}} & \multicolumn{1}{l|}{\textbf{0.860}} & \multicolumn{1}{l|}{\textbf{0.736}} \\ \hline
\multicolumn{1}{|l|}{\# of F1 Win}          & \multicolumn{3}{c|}{0}                                                                       & \multicolumn{3}{c|}{3}                                                                             & \multicolumn{3}{c|}{\textbf{14}}                                                                                \\ \hline
\multicolumn{1}{|l|}{p-value (F1-score)}               & \multicolumn{3}{c|}{$2.7\times 10^{-4}$}                                                                     & \multicolumn{3}{c|}{0.0075}                                                                        & \multicolumn{3}{l|}{}                                                                                           \\ \hline
 &  &  &   &                             &                                &                                     &                                     &                                     & \multicolumn{1}{c}{}               
\end{tabular}}
\vspace{-3mm}
\end{table}


%





\begin{table}[t]
\caption{F1-Score Performance vs. Baselines (without Ranking)}
\vspace{-2mm}
\scalebox{0.55}{
\begin{tabular}{llllllllll}
\cline{2-10}
\multicolumn{1}{l|}{}                       & \multicolumn{3}{c|}{TSC17}                                                                   & \multicolumn{3}{c|}{TSC20}                                                                         & \multicolumn{3}{c|}{Ours (TSC22)}                                                                                          \\ \hline
\multicolumn{1}{|l|}{Pattern Source}        & \multicolumn{1}{c|}{recall} & \multicolumn{1}{c|}{precision} & \multicolumn{1}{c|}{f1-score} & \multicolumn{1}{c|}{recall} & \multicolumn{1}{c|}{precision} & \multicolumn{1}{c|}{f1-score}       & \multicolumn{1}{c|}{recall}          & \multicolumn{1}{c|}{precision}       & \multicolumn{1}{c|}{f1-score}        \\ \hline
\multicolumn{1}{|l|}{Ch. Con.} & \multicolumn{1}{l|}{0.433}  & \multicolumn{1}{l|}{0.882}     & \multicolumn{1}{l|}{0.564}    & \multicolumn{1}{l|}{0.487}  & \multicolumn{1}{l|}{0.648}     & \multicolumn{1}{l|}{0.548}          & \multicolumn{1}{l|}{0.523}           & \multicolumn{1}{l|}{0.920}           & \multicolumn{1}{l|}{\textbf{0.646}}  \\ \hline
\multicolumn{1}{|l|}{BME}                   & \multicolumn{1}{l|}{0.371}  & \multicolumn{1}{l|}{0.836}     & \multicolumn{1}{l|}{0.504}    & \multicolumn{1}{l|}{0.504}  & \multicolumn{1}{l|}{0.719}     & \multicolumn{1}{l|}{0.582}          & \multicolumn{1}{l|}{0.524}           & \multicolumn{1}{l|}{0.933}           & \multicolumn{1}{l|}{\textbf{0.658}}  \\ \hline
\multicolumn{1}{|l|}{TwoPatterns}           & \multicolumn{1}{l|}{0.462}  & \multicolumn{1}{l|}{0.941}     & \multicolumn{1}{l|}{0.597}    & \multicolumn{1}{l|}{0.386}  & \multicolumn{1}{l|}{0.495}     & \multicolumn{1}{l|}{0.421}          & \multicolumn{1}{l|}{0.632}           & \multicolumn{1}{l|}{0.907}           & \multicolumn{1}{l|}{\textbf{0.725}}  \\ \hline
\multicolumn{1}{|l|}{ECG5000}               & \multicolumn{1}{l|}{0.162}  & \multicolumn{1}{l|}{0.583}     & \multicolumn{1}{l|}{0.251}    & \multicolumn{1}{l|}{0.748}  & \multicolumn{1}{l|}{0.853}     & \multicolumn{1}{l|}{0.787}          & \multicolumn{1}{l|}{0.783}           & \multicolumn{1}{l|}{1.000}           & \multicolumn{1}{l|}{\textbf{0.875}}  \\ \hline
\multicolumn{1}{|l|}{CBF}                   & \multicolumn{1}{l|}{0.183}  & \multicolumn{1}{l|}{0.611}     & \multicolumn{1}{l|}{0.278}    & \multicolumn{1}{l|}{0.353}  & \multicolumn{1}{l|}{0.556}     & \multicolumn{1}{l|}{0.419}          & \multicolumn{1}{l|}{0.685}           & \multicolumn{1}{l|}{0.947}           & \multicolumn{1}{l|}{\textbf{0.782}}  \\ \hline
\multicolumn{1}{|l|}{TwoLeadECG}            & \multicolumn{1}{l|}{0.161}  & \multicolumn{1}{l|}{0.547}     & \multicolumn{1}{l|}{0.246}    & \multicolumn{1}{l|}{0.834}  & \multicolumn{1}{l|}{0.894}     & \multicolumn{1}{l|}{\textbf{0.850}} & \multicolumn{1}{l|}{0.683}           & \multicolumn{1}{l|}{0.967}           & \multicolumn{1}{l|}{0.789}           \\ \hline
\multicolumn{1}{|l|}{ECG200}                & \multicolumn{1}{l|}{0.183}  & \multicolumn{1}{l|}{0.650}     & \multicolumn{1}{l|}{0.283}    & \multicolumn{1}{l|}{0.511}  & \multicolumn{1}{l|}{0.808}     & \multicolumn{1}{l|}{\textbf{0.617}} & \multicolumn{1}{l|}{0.409}           & \multicolumn{1}{l|}{0.900}           & \multicolumn{1}{l|}{0.550}           \\ \hline
\multicolumn{1}{|l|}{ECGFiveDays}           & \multicolumn{1}{l|}{0.180}  & \multicolumn{1}{l|}{0.650}     & \multicolumn{1}{l|}{0.279}    & \multicolumn{1}{l|}{0.586}  & \multicolumn{1}{l|}{0.660}     & \multicolumn{1}{l|}{0.609}          & \multicolumn{1}{l|}{0.738}           & \multicolumn{1}{l|}{0.992}           & \multicolumn{1}{l|}{\textbf{0.841}}  \\ \hline
\multicolumn{1}{|l|}{FreezerSmallTrain}     & \multicolumn{1}{l|}{0.229}  & \multicolumn{1}{l|}{0.781}     & \multicolumn{1}{l|}{0.350}    & \multicolumn{1}{l|}{0.710}  & \multicolumn{1}{l|}{0.790}     & \multicolumn{1}{l|}{0.734}          & \multicolumn{1}{l|}{0.787}           & \multicolumn{1}{l|}{1.000}           & \multicolumn{1}{l|}{\textbf{0.873}}  \\ \hline
\multicolumn{1}{|l|}{RegularSmallTrain}     & \multicolumn{1}{l|}{0.256}  & \multicolumn{1}{l|}{0.844}     & \multicolumn{1}{l|}{0.388}    & \multicolumn{1}{l|}{0.647}  & \multicolumn{1}{l|}{0.782}     & \multicolumn{1}{l|}{0.686}          & \multicolumn{1}{l|}{0.782}           & \multicolumn{1}{l|}{0.954}           & \multicolumn{1}{l|}{\textbf{0.844}}  \\ \hline
\multicolumn{1}{|l|}{Trace}                 & \multicolumn{1}{l|}{0.225}  & \multicolumn{1}{l|}{0.783}     & \multicolumn{1}{l|}{0.346}    & \multicolumn{1}{l|}{0.691}  & \multicolumn{1}{l|}{0.837}     & \multicolumn{1}{l|}{0.744}          & \multicolumn{1}{l|}{0.677}           & \multicolumn{1}{l|}{0.967}           & \multicolumn{1}{l|}{\textbf{0.772}}  \\ \hline
\multicolumn{1}{|l|}{Wafer}                 & \multicolumn{1}{l|}{0.317}  & \multicolumn{1}{l|}{0.728}     & \multicolumn{1}{l|}{0.421}    & \multicolumn{1}{l|}{0.516}  & \multicolumn{1}{l|}{0.675}     & \multicolumn{1}{l|}{0.571}          & \multicolumn{1}{l|}{0.607}           & \multicolumn{1}{l|}{0.900}           & \multicolumn{1}{l|}{\textbf{0.705}}  \\ \hline
\multicolumn{1}{|l|}{Plane}                 & \multicolumn{1}{l|}{0.161}  & \multicolumn{1}{l|}{0.547}     & \multicolumn{1}{l|}{0.246}    & \multicolumn{1}{l|}{0.832}  & \multicolumn{1}{l|}{0.892}     & \multicolumn{1}{l|}{0.859}          & \multicolumn{1}{l|}{0.884}           & \multicolumn{1}{l|}{1.000}           & \multicolumn{1}{l|}{\textbf{0.932}}  \\ \hline
\multicolumn{1}{|l|}{SonyAI}                & \multicolumn{1}{l|}{0.158}  & \multicolumn{1}{l|}{0.556}     & \multicolumn{1}{l|}{0.244}    & \multicolumn{1}{l|}{0.341}  & \multicolumn{1}{l|}{0.500}     & \multicolumn{1}{l|}{0.391}          & \multicolumn{1}{l|}{0.544}           & \multicolumn{1}{l|}{0.933}           & \multicolumn{1}{l|}{\textbf{0.675}}  \\ \hline
\multicolumn{1}{|l|}{SonyAI2}               & \multicolumn{1}{l|}{0.160}  & \multicolumn{1}{l|}{0.563}     & \multicolumn{1}{l|}{0.247}    & \multicolumn{1}{l|}{0.333}  & \multicolumn{1}{l|}{0.568}     & \multicolumn{1}{l|}{0.407}          & \multicolumn{1}{l|}{0.479}           & \multicolumn{1}{l|}{0.847}           & \multicolumn{1}{l|}{\textbf{0.603}}  \\ \hline
\multicolumn{1}{|l|}{Lightning7}            & \multicolumn{1}{l|}{0.216}  & \multicolumn{1}{l|}{0.800}     & \multicolumn{1}{l|}{0.337}    & \multicolumn{1}{l|}{0.336}  & \multicolumn{1}{l|}{0.610}     & \multicolumn{1}{l|}{0.428}          & \multicolumn{1}{l|}{0.502}           & \multicolumn{1}{l|}{0.916}           & \multicolumn{1}{l|}{\textbf{0.636}}  \\ \hline
\multicolumn{1}{|l|}{UME}                   & \multicolumn{1}{l|}{0.480}  & \multicolumn{1}{l|}{0.933}     & \multicolumn{1}{l|}{0.623}    & \multicolumn{1}{l|}{0.531}  & \multicolumn{1}{l|}{0.745}     & \multicolumn{1}{l|}{0.616}          & \multicolumn{1}{l|}{0.517}           & \multicolumn{1}{l|}{0.943}           & \multicolumn{1}{l|}{\textbf{0.657}}  \\ \hline
\multicolumn{1}{|l|}{Average}               & \multicolumn{1}{l|}{0.2551} & \multicolumn{1}{l|}{0.7198}    & \multicolumn{1}{l|}{0.3649}   & \multicolumn{1}{l|}{0.5498} & \multicolumn{1}{l|}{0.7077}    & \multicolumn{1}{l|}{0.6041}         & \multicolumn{1}{l|}{\textbf{0.6327}} & \multicolumn{1}{l|}{\textbf{0.9427}} & \multicolumn{1}{l|}{\textbf{0.7390}} \\ \hline
\multicolumn{1}{|l|}{\# of F1 Win}          & \multicolumn{3}{c|}{0}                                                                      & \multicolumn{3}{c|}{2}                                                                             & \multicolumn{3}{c|}{\textbf{15}}                                                                                   \\ \hline
\multicolumn{1}{|l|}{p-value (F1-score)}               & \multicolumn{3}{c|}{$2.7\times 10^{-4}$}                                                                     & \multicolumn{3}{c|}{0.001}                                                                         & \multicolumn{3}{l|}{}                                                                                              \\ \hline
                                            &                             &                                &                               &                             &                                &                                     &                                      &                                      & \multicolumn{1}{c}{}                
\end{tabular}}
\vspace{-5mm}
\end{table}

\subsubsection{Overall Performance with Ranking}
 
 Table I shows the performance of the top-1 chain detected by all three methods. TSC22 outperforms both baselines on 14 out of 17 datasets in terms of the F1-score, and the average (0.736) is much better than that of TSC17 (0.225) and TSC20 (0.604).
 The Wilcoxon signed-rank test p-value between TSC22 and TSC17 is $2.7\times10^{-4}$ and that between TSC22 and TSC20 is 0.0075. The fact that both p-values are less than 0.05 indicate that TSC22 outperforms the baselines with statistical significance.
\subsubsection{Performance without Ranking}
To fairly compare all three TSC methods on the definitions without the effect of ranking, we evaluated the performance on the discovered chains by fixing the start node of the chain. In each time series we ``grew" chains from the last 5 nodes in the ground truth chain independently, and reported the maximum F1-score obtained among these 5 different chains.
From Table II, we can see that TSC22 outperforms both baselines on 15 out of 17 datasets in term of F1-score. As the p-values of the Wilcoxon tests are less than 0.05, we conclude that the TSC22 definition significantly outperforms the existing TSC definitions.

\vspace{-1mm}
\section{Conclusion}
In this work, we propose a novel time series chain definition $TSC22$, which exploits an idea to track how the nearest neighbor of a pattern changes over time to improve the robustness of the chain against noise. In addition, two new quality metrics are proposed to effectively rank the detect chains. Extensive experiments on both real-world and synthetic data show that our new definition is much more robust than the state-of-the-art TSC definitions. The case studies conducted on real-world time series also show that the top ranked chain discovered by our method can reveal meaningful regularities in a variety of applications under different noisy conditions.



\bibliographystyle{IEEEtran}
\bibliography{TSC}

\begin{thebibliography}{10}
\providecommand{\url}[1]{#1}
\csname url@samestyle\endcsname
\providecommand{\newblock}{\relax}
\providecommand{\bibinfo}[2]{#2}
\providecommand{\BIBentrySTDinterwordspacing}{\spaceskip=0pt\relax}
\providecommand{\BIBentryALTinterwordstretchfactor}{4}
\providecommand{\BIBentryALTinterwordspacing}{\spaceskip=\fontdimen2\font plus
\BIBentryALTinterwordstretchfactor\fontdimen3\font minus
  \fontdimen4\font\relax}
\providecommand{\BIBforeignlanguage}[2]{{%
\expandafter\ifx\csname l@#1\endcsname\relax
\typeout{** WARNING: IEEEtran.bst: No hyphenation pattern has been}%
\typeout{** loaded for the language `#1'. Using the pattern for}%
\typeout{** the default language instead.}%
\else
\language=\csname l@#1\endcsname
\fi
#2}}
\providecommand{\BIBdecl}{\relax}
\BIBdecl

\bibitem{zhumatrix}
Y.~Zhu, Z.~Zimmerman, N.~S. Senobari, C.-C.~M. Yeh, G.~Funning, A.~Mueen,
  P.~Brisk, and E.~Keogh, ``Matrix profile ii: Exploiting a novel algorithm and
  gpus to break the one hundred million barrier for time series motifs and
  joins,'' in \emph{Data Mining (ICDM), 2016 IEEE 16th International Conference
  on}.\hskip 1em plus 0.5em minus 0.4em\relax IEEE, 2016, pp. 739--748.

\bibitem{zhang2020semantic}
L.~Zhang, Y.~Gao, and J.~Lin, ``Semantic discord: Finding unusual local
  patterns for time series,'' in \emph{Proceedings of the 2020 SIAM
  International Conference on Data Mining}.\hskip 1em plus 0.5em minus
  0.4em\relax SIAM, 2020, pp. 136--144.

\bibitem{yeh2016matrix}
C.-C.~M. Yeh, Y.~Zhu, L.~Ulanova, N.~Begum, Y.~Ding, H.~A. Dau, D.~F. Silva,
  A.~Mueen, and E.~Keogh, ``Matrix profile i: all pairs similarity joins for
  time series: a unifying view that includes motifs, discords and shapelets,''
  in \emph{2016 IEEE 16th international conference on data mining
  (ICDM)}.\hskip 1em plus 0.5em minus 0.4em\relax Ieee, 2016, pp. 1317--1322.

\bibitem{gao2017trajviz}
Y.~Gao, Q.~Li, X.~Li, J.~Lin, and H.~Rangwala, ``Trajviz: a tool for
  visualizing patterns and anomalies in trajectory,'' in
  \emph{ECML-PKDD}.\hskip 1em plus 0.5em minus 0.4em\relax Springer, 2017, pp.
  428--431.

\bibitem{zhu2017matrix}
Y.~Zhu, M.~Imamura, D.~Nikovski, and E.~Keogh, ``Matrix profile vii: Time
  series chains: A new primitive for time series data mining (best student
  paper award),'' in \emph{2017 IEEE International Conference on Data Mining
  (ICDM)}.\hskip 1em plus 0.5em minus 0.4em\relax IEEE, 2017, pp. 695--704.

\bibitem{imamura2020matrix}
M.~Imamura, T.~Nakamura, and E.~Keogh, ``Matrix profile xxi: A geometric
  approach to time series chains improves robustness,'' in \emph{Proceedings of
  the 26th ACM SIGKDD International Conference on Knowledge Discovery \& Data
  Mining}, 2020, pp. 1114--1122.

\bibitem{gao2018exploring}
Y.~Gao and J.~Lin, ``Exploring variable-length time series motifs in one
  hundred million length scale,'' \emph{Data Mining and Knowledge Discovery},
  vol.~32, no.~5, pp. 1200--1228, 2018.

\bibitem{keogh2005hot}
E.~Keogh, J.~Lin, and A.~Fu, ``Hot sax: Efficiently finding the most unusual
  time series subsequence,'' in \emph{Fifth IEEE International Conference on
  Data Mining (ICDM'05)}.\hskip 1em plus 0.5em minus 0.4em\relax IEEE, 2005,
  pp. 8--pp.

\bibitem{senin2015time}
P.~Senin, J.~Lin, X.~Wang, T.~Oates, S.~Gandhi, A.~P. Boedihardjo, C.~Chen, and
  S.~Frankenstein, ``Time series anomaly discovery with grammar-based
  compression.'' in \emph{Edbt}, 2015, pp. 481--492.

\bibitem{paparrizos2015k}
J.~Paparrizos and L.~Gravano, ``k-shape: Efficient and accurate clustering of
  time series,'' in \emph{Proceedings of the 2015 ACM SIGMOD international
  conference on management of data}, 2015, pp. 1855--1870.

\bibitem{ma2019learning}
Q.~Ma, J.~Zheng, S.~Li, and G.~W. Cottrell, ``Learning representations for time
  series clustering,'' \emph{Advances in neural information processing
  systems}, vol.~32, 2019.

\bibitem{li2021time}
X.~Li, J.~Lin, and L.~Zhao, ``Time series clustering in linear time
  complexity,'' \emph{Data Mining and Knowledge Discovery}, vol.~35, no.~6, pp.
  2369--2388, 2021.

\bibitem{anton2018time}
S.~D. Anton, L.~Ahrens, D.~Fraunholz, and H.~D. Schotten, ``Time is of the
  essence: Machine learning-based intrusion detection in industrial time series
  data,'' in \emph{2018 IEEE International Conference on Data Mining Workshops
  (ICDMW)}.\hskip 1em plus 0.5em minus 0.4em\relax IEEE, 2018, pp. 1--6.

\bibitem{bloch1997practical}
H.~P. Bloch and F.~K. Geitner, \emph{Practical Machinery Management for Process
  Plants: Volume 2: Machinery Failure Analysis and Troubleshooting}.\hskip 1em
  plus 0.5em minus 0.4em\relax Elsevier, 1997.

\bibitem{ahmad2012overview}
R.~Ahmad and S.~Kamaruddin, ``An overview of time-based and condition-based
  maintenance in industrial application,'' \emph{Computers \& industrial
  engineering}, vol.~63, no.~1, pp. 135--149, 2012.

\bibitem{jardine2006review}
A.~K. Jardine, D.~Lin, and D.~Banjevic, ``A review on machinery diagnostics and
  prognostics implementing condition-based maintenance,'' \emph{Mechanical
  systems and signal processing}, vol.~20, no.~7, pp. 1483--1510, 2006.

\bibitem{wang2019discovering}
S.~Wang, Y.~Yuan, and H.~Li, ``Discovering all-chain set in streaming time
  series,'' in \emph{PAKDD}.\hskip 1em plus 0.5em minus 0.4em\relax Springer,
  2019, pp. 306--318.

\bibitem{zhang2022joint}
L.~Zhang, N.~Patel, X.~Li, and J.~Lin, ``Joint time series chain: Detecting
  unusual evolving trend across time series,'' in \emph{Proceedings of the 2022
  SIAM International Conference on Data Mining (SDM)}.\hskip 1em plus 0.5em
  minus 0.4em\relax SIAM, 2022, pp. 208--216.

\bibitem{lin2007experiencing}
J.~Lin, E.~Keogh, L.~Wei, and S.~Lonardi, ``Experiencing sax: a novel symbolic
  representation of time series,'' \emph{Data Mining and knowledge discovery},
  vol.~15, no.~2, pp. 107--144, 2007.

\bibitem{zhu2020index}
Y.~Zhu, C.-C.~M. Yeh, Z.~Zimmerman, and E.~Keogh, ``Matrix profile xvii:
  Indexing the matrix profile to allow arbitrary range queries,'' in \emph{2020
  IEEE 36th International Conference on Data Engineering (ICDE)}, 2020, pp.
  1846--1849.

\bibitem{supportingwebpage22}
\BIBentryALTinterwordspacing
Supporting webpage. [Online]. Available:
  \url{https://sites.google.com/view/robust-time-series-chain-22}
\BIBentrySTDinterwordspacing

\bibitem{wu2021current}
R.~Wu and E.~Keogh, ``Current time series anomaly detection benchmarks are
  flawed and are creating the illusion of progress,'' \emph{IEEE Transactions
  on Knowledge and Data Engineering}, 2021.

\end{thebibliography}

\end{document}